\documentclass{article} 
\usepackage{iclr2027_conference,times}

\usepackage{amsmath,amsfonts,bm}

\def\eqref#1{equation~\ref{#1}}

\def\1{\bm{1}}

\DeclareMathAlphabet{\mathsfit}{\encodingdefault}{\sfdefault}{m}{sl}
\SetMathAlphabet{\mathsfit}{bold}{\encodingdefault}{\sfdefault}{bx}{n}

\usepackage{graphicx}
\usepackage{hyperref}
\usepackage{booktabs}
\usepackage{array}
\usepackage{colortbl}
\usepackage{capt-of}
\hypersetup{
 colorlinks=true,
 linkcolor=red,
 citecolor=cyan,
 filecolor=magenta, 
 urlcolor=red,
 }
\usepackage{url}
\usepackage[utf8]{inputenc}
\usepackage[T1]{fontenc}
\usepackage{fvextra}
\usepackage{tcolorbox}
\tcbuselibrary{breakable}

\DeclareUnicodeCharacter{2014}{\textemdash}
\DeclareUnicodeCharacter{2026}{\textellipsis}
\DeclareUnicodeCharacter{2192}{\ensuremath{\rightarrow}}

\tcbset{
    myframe/.style={
        breakable,
        colback=#1!5!white,
        colframe=#1!65!black,
        colbacktitle=#1!15!white,
        coltitle=black,
        fonttitle=\bfseries\small,
        boxrule=0.5pt,
        arc=1mm,
        left=2mm,
        right=2mm,
        top=2mm,
        bottom=2mm
    }
}

\title{
Agentic-TTT: Training test-time policy \\for test-time training}

\author{Jiahao Lu \qquad Mohan Kankanhalli \\
NUS AI Institute\\
National University of Singapore\\
\texttt{jiahao.lu@u.nus.edu \quad dcsmsk@nus.edu.sg} \\
}

\iclrfinalcopy 
\begin{document}

\maketitle

\begin{abstract}

Test-time training (TTT) adapts an LLM's parameters using signals derived from test inputs, 
and can make striking improvements in pre-specified settings such as IMO competitions or designated open problems.
By turning deployment experience into parameter updates, 
TTT provides a direct mechanism for model-level self-improvement.
Yet TTT is not universally beneficial:
each TTT algorithm works in different settings,
and applying an ill-suited method could waste test-time compute or even damage model performance.
Therefore, such parameter-level self-improvement requires agency:
the model must decide 
when TTT is warranted,
which algorithm to invoke,
and whether an existing skill can be reused.
To fill this gap, we introduce \emph{Agentic-TTT},
which learns a test-time policy to govern those decisions.
Agentic-TTT turns TTT procedures into callable tools,
treats accumulated skills as an evolving deployment environment, 
and trains its policy using the observed utility gains from its decisions.
On our benchmark,
Agentic-TTT nearly doubles the utility over the backbone model,
learns to trade off utility against compute,
and generalizes to domains unseen during training.
Together, these results point toward autonomous self-improvement:
models that can decide how to learn from their own deployment experience.
\end{abstract}
\section{Introduction}
\label{sec:introduction}

In 1945, von Neumann proposed the stored-program architecture, enabling \emph{self-modifying code}:
programs that rewrite their own instructions based on runtime data~\citep{vonneumann1945first}.
Alan Turing later recognized a deeper promise in such self-modification, arguing that 
``the possibility of letting the machine alter its own instructions provides the mechanism for learning from experience''~\citep{turing1947lecture}.
Eight decades later, neural networks have acquired remarkable abilities to perceive, generate, and reason, 
yet their parameters are typically frozen after deployment.
Data encountered during deployment can shape the model's computation, 
but ordinarily leaves its parameters and thus its parameterized capabilities unchanged.
What would it take for a neural model to continue growing its own capabilities after deployment?

A prominent line of modern self-improving agents carries this vision forward at the level of the surrounding system.
Through reflection and memory, these agents distill past interactions into textual lessons that guide subsequent decisions~\citep{shinn2023reflexion,zhao2024expel}.
They can also evolve their prompts, search over entire agentic workflows, 
and accumulate executable skills for reuse across tasks~\citep{fernando2024promptbreeder,zhang2025aflow,wang2024voyager}. 
Although these mechanisms genuinely improve the agent as a whole, the self-improvement largely resides in the scaffold around the model,
including its context, control flow, and external action repertoire; the model parameters themselves remain unchanged.

Other self-evolving systems update model parameters through self-generated tasks and trajectories,
but organize this learning before deployment~\citep{hu2025agentgen, zhao2026absolute}.
Test-time training (TTT) moves parameter adaptation into deployment itself, 
using signals derived from test inputs to update model weights~\citep{zuo2025ttrl,zweiger2025seal,akyurek2025fsl,yuksekgonul2026discover},
enabling model-level self-improvement from deployment experience.
TTT has achieved striking results in pre-specified settings, 
including AlphaProof's silver-medal IMO performance~\citep{hubert2025olympiad} and TTT-Discover's advances on open scientific problems~\citep{yuksekgonul2026discover}.


Yet these methods typically assume the deployer has already identified both the capability to acquire and the procedure for acquiring it. 
In open-world deployment, where the test distribution is heterogeneous and unpredictable, 
an ill-suited TTT procedure can waste substantial test-time compute and may even underperform the frozen model.
Realizing deployment-time capability growth therefore requires \textbf{agency}:
deciding whether a query warrants adaptation, whether a capability acquired earlier already suffices, 
and which TTT method can best convert test-time compute into utility gain.
To our knowledge, this model-level agency over TTT remains unexplored in open-ended LLM deployment.
We fill this gap with \emph{Agentic-TTT}, which turns TTT procedures into model-invokable tools and trains a test-time policy to govern their use.

Agentic-TTT formulates this agency as sequential decision-making over a stream of test queries.
At a high level, given a query and the current skill library, the policy decides whether to answer with the frozen model, reuse an existing skill, or acquire a new capability through a selected TTT method.
Each TTT invocation produces an isolated LoRA skill over the frozen backbone~\citep{hu2022lora}.
Storing these skills in a persistent library expands the options available to future queries, 
making the skill library an evolving capability environment.
We train the policy to maximize utility gain over the frozen model while accounting for test-time computation.
Agentic-TTT thereby turns the model from a passive target of adaptation into an active agent of self-improvement.


We evaluate Agentic-TTT on a heterogeneous query stream
drawn from eleven benchmark families across eight domains.
We deliberately enrich the benchmark so that most queries
exhibit TTT benefits during screening,
allowing us to compare how effectively different policies
exploit these adaptation opportunities.
The reported gains therefore characterize this curated evaluation setting.
The evaluated fixed-method baselines show uneven benefits
and can degrade performance in some domains.
In contrast, Agentic-TTT nearly doubles
Qwen2.5-7B's overall utility over direct inference,
with positive gains in every evaluated domain.
The policy also supports controllable utility-compute trade-offs
and retains substantial gains on domains excluded from policy training.
Improvements persist with the stronger Qwen2.5-14B backbone.

Taken together, our work reframes test-time training from a fixed adaptation procedure into a problem of learned agency over deployment-time self-improvement. 
Agentic-TTT realizes this agency through a test-time policy that governs when and how to build or reuse parameter-space skills during deployment.
Across a diverse suite of test domains,
the learned policy nearly doubles overall utility over direct inference,
delivers positive gains in every evaluated domain,
supports controllable utility-compute trade-offs,
and generalizes its TTT decisions to unseen data sources.
These results suggest that the next frontier of TTT lies not in a single universal adaptation rule, but in models that learn when and how to draw on a diverse portfolio of TTT methods for self-improvement.

\begin{figure}[h!]
    \centering
    \includegraphics[width=0.8\textwidth]{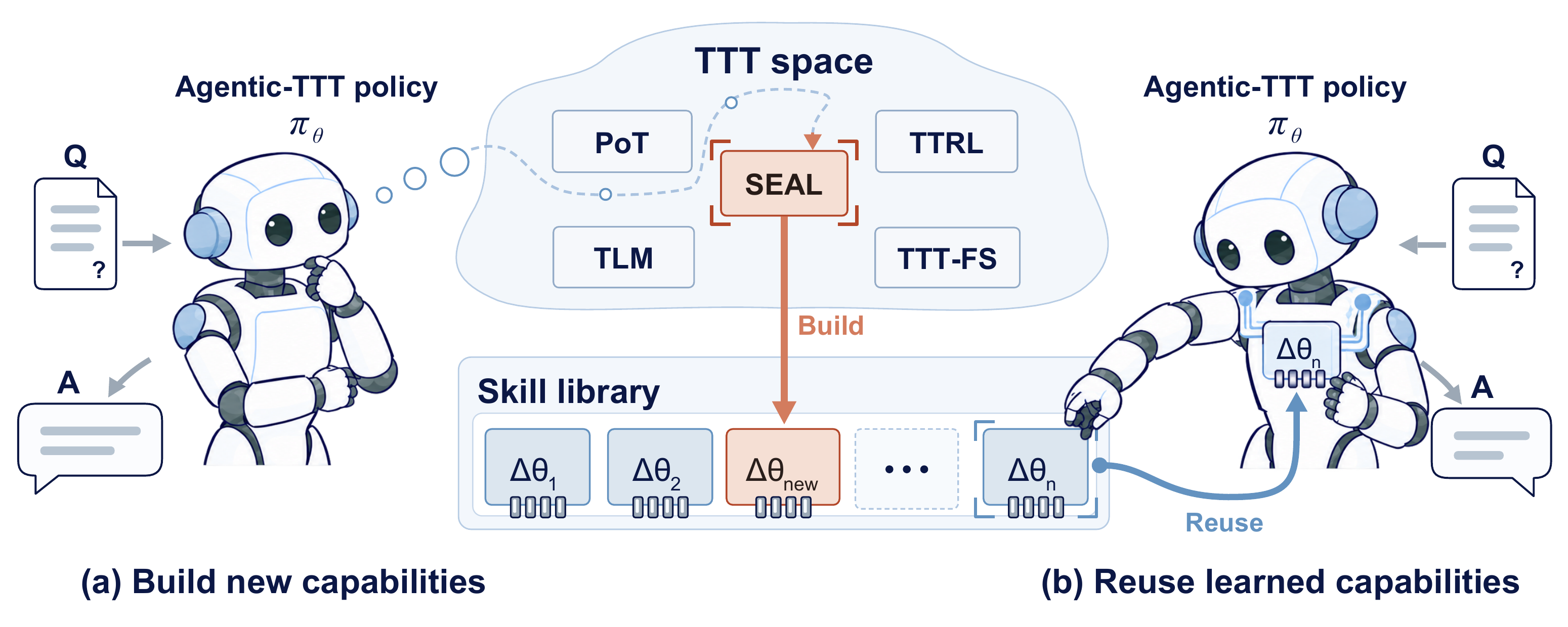}
    \caption{\textbf{Agentic-TTT learns to govern self-improvement at test time:} a learned policy decides when and how to (a) build or (b) reuse parameterized skills through test-time training.}
    \label{fig:agentic-ttt-teaser}
\end{figure}

\section{Agentic Test-Time Training}
\label{sec:agentic_ttt}

\subsection{Problem Formulation}
\label{sec:problem_formulation}

Test-time training (TTT) offers the potential for self-improvement at deployment, 
but realizing such potential requires knowing which adaptation, if any, is appropriate for each test query.
Agentic-TTT provides this missing agency: 
it learns when and how to build new capabilities through TTT, and when to improve an answer by creating or reusing TTT skills.

We formalize this agency as sequential decision-making over a stream of test queries \(x_1,\ldots,x_T\). 
Some TTT methods can adapt from a single query when it contains sufficient training material.
For example, 
SEAL~\citep{zweiger2025seal} learns from a bundled document, whereas 
TTT-fewshot~\citep{akyurek2025fsl} adapts from few-shot demonstrations.
By contrast, 
a multi-query method such as TTRL~\citep{zuo2025ttrl}
aggregates self-consistency signals across multiple same-domain queries.
To support such methods,
the environment maintains adaptation pools \(\mathcal P_t\).
Each pool \(\rho \in \mathcal P_t\) stores a collection of compatible queries,
together with any bundled material needed for adaptation,
until they can be used jointly in a TTT run.
The base model remains frozen throughout.
Whether produced from a single query or an adaptation pool,
each adaptation is represented as a LoRA adapter
and stored as a reusable parametric skill \(s\) in the skill library \(\mathcal L_t\).
Each skill \(s\) is accompanied by a natural-language description of its learned capability, 
allowing the policy to assess its relevance to future queries.
The pools and library persist across queries and shape subsequent decisions,
forming the cross-query state
\(S_t=\left(\mathcal L_t,\mathcal P_t\right)\)
before processing each \(x_t\).
Processing \(x_t\) may update either adaptation pools or skill library, and the resulting evolved state  \(S_{t+1}\) is carried forward to \(x_{t+1}\).

Processing a query may require a sequence of actions over multiple interaction turns.
At each turn \(k\),
the policy selects one of four primitive actions:
\begin{equation}
    a_{t,k}
    \in
    \mathcal A
    =
    \left\{
        \textsc{Assign}(\rho),
        \textsc{Adapt}(m,\rho),
        \textsc{Load}(s),
        \textsc{Generate}
    \right\},
    \label{eq:agentic_action}
\end{equation}

where \textsc{Assign} adds adaptation material from \(x_t\) to pool \(\rho\),
\textsc{Adapt} applies TTT method \(m\) to pool \(\rho\) and produces a skill,
\textsc{Load} mounts an existing skill \( s \) ,
whereas \textsc{Generate} produces the final answer and terminates the interaction.
To select \(a_{t,k}\), 
the policy conditions on the current query,
the cross-query state,
and the preceding interaction history.
We denote this context by \(c_{t,k}\).
The policy first generates a reasoning trace \(z_{t,k}\)
and then selects the action \(a_{t,k}\) conditioned on that reasoning.
We denote the complete sequence of reasoning traces,
actions,
and environment feedback by \(\tau_t\),
and its prefix before turn \(k\) by \(\tau_{t,<k}\).
After executing \(a_{t,k}\),
the environment appends its feedback to \(\tau_{t,<k+1} \). 
Formally,

\begin{equation}
    c_{t,k}
    =
    \left(
        x_t,
        S_t,
        \tau_{t,<k}
    \right),
    \qquad
    z_{t,k}
    \sim
    \pi_\theta
    \left(
        \cdot
        \mid
        c_{t,k}
    \right),
    \qquad
    a_{t,k}
    \sim
    \pi_\theta
    \left(
        \cdot
        \mid
        c_{t,k},
        z_{t,k}
    \right)
    .
    \label{eq:agentic_policy}
\end{equation}

Let \(U(\tau_t)\) denote the utility of the final answer, 
and let \(C(\tau_t)\) denote the computation cost modeling of the trajectory, which may include policy reasoning, TTT actions and final answer generation. 
We compare each trajectory with direct inference with the frozen base model \(\tau_t^{\textsc{Generate}}\), and define the base-relative return as:
\begin{equation}
    R(\tau_t)
    =
    \left[
        U(\tau_t)
        -
        U\!\left(
            \tau_t^{\textsc{Generate}}
        \right)
    \right]
    -
    \lambda
    \left[
        C(\tau_t)
        -
        C\!\left(
            \tau_t^{\textsc{Generate}}
        \right)
    \right],
    \label{eq:agentic_reward}
\end{equation}

where \(\lambda\geq 0\) controls the trade-off
between answer quality and computation. 
An agentic trajectory is therefore beneficial
only when its utility improvement justifies its additional cost.
Agentic-TTT therefore seeks a policy \(\pi_{\theta^\star}\) that 
maximizes the expected cumulative return when deployed over a query stream:

\vspace{-1em}
\begin{equation}
    \theta^\star
    =
    \arg\max_\theta
    \mathbb E_{\pi_\theta}
    \left[
        \sum_{t=1}^{T}
        R(\tau_t)
    \right].
    \label{eq:stream_objective}
\end{equation}


\subsection{Learning the Agentic-TTT Policy}
\label{sec:policy_learning}

Learning the Agentic-TTT policy directly from outcome rewards poses a cold-start problem.
Since the initial policy is unfamiliar with the execution protocol,
it may rarely sample valid trajectories, 
leaving little useful signal for reinforcement learning. 
We therefore train the policy in two stages.
\textit{Guidance distillation policy learning} first establishes reliable interaction and a useful prior over TTT decisions.
Building on this behavioral foundation,
\textit{outcome-driven reinforcement learning} further maximizes the measured utility gain of each trajectory while penalizing its computation cost.

\subsubsection{Guidance Distillation Policy Learning}
\label{sec:guidance_distillation}

Since the base model has never been exposed to our execution harness,
appropriate action sequences may initially have negligible probability. 
When the policy is unlikely to sample a valid action sequence, 
the correct behavior remains almost invisible to reward-based learning and cannot be effectively reinforced. 
We address this cold-start issue by \textit{Guidance Distillation Policy Learning}.

We collect SFT demonstrations through hint-guided rollout collection.
For each training query \(x_t\),
heuristic rules propose a sequence of turn-level gold actions \(a_{t,k}^{\star}\),
which we retain only after empirical TTT evaluation confirms its predicted benefit.
At each turn,
a context-specific hint \(h_{t,k}\) identifies the next gold action and explains why it is appropriate.
The initial policy \(\pi_{\theta_0}\) then samples its own reasoning and action:
\begin{equation}
    (z_{t,k},a_{t,k})
    \sim
    \pi_{\theta_0}
    \left(
        \cdot
        \mid
        c_{t,k},
        h_{t,k}
    \right),
    \qquad
    \text{retain only if }
    a_{t,k}=a_{t,k}^{\star}.
    \label{eq:guidance_collection}
\end{equation}

After each accepted action is executed,
its environment feedback enters \(c_{t,k+1}\),
and a new hint guides the next turn until the trajectory terminates.
Each rationale \(z_{t,k}\) is thus generated on-policy by \(\pi_{\theta_0}\) itself,
rather than imposing an external teacher's reasoning style.

After collection,
we remove all hints and construct \(\mathcal D_{\mathrm{SFT}}\) from all the turn-level triples \((c_{t,k},z_{t,k},a_{t,k}^{\star})\). 
Tokens in \(c_{t,k}\) serve only as the conditioning context and are masked from the loss,
so supervision applies only to the self-generated rationale \(z_{t,k}\) and the gold action block \(a_{t,k}^{\star}\).
The action block \(a_{t,k}^{\star}\) must follow a canonical harness-defined format that is unfamiliar to the initial policy yet crucial for reliable execution.
Under a uniform token-averaged objective,
longer rationales would dilute the learning signal assigned to the action tokens;
therefore, we optimize the following role-weighted SFT objective:
\begin{equation}
    \mathcal L_{\mathrm{SFT}}(\theta)
    =
    -
    \mathbb E_{\mathcal D_{\mathrm{SFT}}}
    \left[
        \omega_z
        \log
        \pi_\theta
        \left(
            z_{t,k}
            \mid
            c_{t,k}
        \right)
        +
        \omega_a
        \log
        \pi_\theta
        \left(
            a_{t,k}^{\star}
            \mid
            c_{t,k},
            z_{t,k}
        \right)
    \right],
    \label{eq:guidance_distillation_objective}
\end{equation}
We set \(\omega_z=1\) and \(\omega_a=4\), 
giving action tokens stronger supervision guidance than rationale tokens.
Minimizing this objective yields \(\pi_{\theta_{\mathrm{SFT}}}\),
which initializes outcome-driven reinforcement learning.

\subsubsection{Outcome-Driven Reinforcement Learning}
\label{sec:outcome_driven_rl}

We initialize \(\pi_\theta\) from \(\pi_{\theta_{\mathrm{SFT}}}\) and refine it through \textit{outcome-driven reinforcement learning}. 
By optimizing multi-turn trajectory outcomes directly,
the policy learns when the utility gain from TTT justifies its cost and when direct generation is preferable.

Since deployment-time self-improvement depends on capabilities accumulated across queries, 
we train the policy in a streaming environment.
When processing \(x_t\),
the policy observes a cross-query state \(S_t\) shaped by the training events that precede it.
At the beginning of each epoch,
we reset the skill library and adaptation pools and reshuffle the training queries subject to prerequisite-order constraints.
This exposes each query to varied cross-query states and reduces overfitting to fixed \((x_t,S_t)\) pairs.
When a training event \((x_t,S_t)\) is reached, 
we snapshot \(S_t\) and independently sample \(G\) complete trajectories:
\[
    \tau_t^{(g)}
    \sim
    \pi_\theta
    \left(
        \cdot
        \mid
        x_t,S_t
    \right),
    \qquad
    g=1,\ldots,G.
    \label{eq:rl_rollouts}
\]

These rollouts provide local counterfactual outcomes from the same state snapshot and are rolled back after evaluation and optimization.
The training stream instead advances by executing the annotated gold action sequence for \(x_t\),
producing a single valid successor state \(S_{t+1}\).
Oracle stream advancement therefore determines only the cross-query states on which RL is trained;
at deployment,
the policy advances the stream autonomously.

For each sampled rollout \(\tau=\tau_t^{(g)}\),
we set the reward as the base-relative return 
\begin{equation}
    \begin{aligned}
        R(\tau)
        &=
        \Delta U(\tau)
        -
        \lambda_{\mathrm{comp}}
        \Delta C(\tau)
        -
        \lambda_{\mathrm{inv}}
        P_{\mathrm{inv}}(\tau),
        \\
        &\text{where}\quad
        \Delta U(\tau)
        :=
        U(\tau)
        -
        U
        \left(
            \tau^{\textsc{Generate}}
        \right),
        \\
        &\phantom{\text{where}\quad}
        \Delta C(\tau)
        :=
        C(\tau)
        -
        C
        \left(
            \tau^{\textsc{Generate}}
        \right).
    \end{aligned}
    \label{eq:outcome_reward}
\end{equation}

Here,
\(\tau^{\textsc{Generate}}\) denotes the direct answer trajectory from the frozen base model without \textsc{Assign}, \textsc{Adapt}, or \textsc{Load} actions. 
The cost model \(C\) and coefficient \(\lambda_{\mathrm{comp}}\) are instantiated according to the desired utility-compute trade-off,
with the exact configurations specified in the experimental setup.
\(P_{\mathrm{inv}}(\tau)\) is a graded penalty for invalid actions.
We observe that the policy may perform invalid actions such as invoking nonexistent skills or attempting adaptation before its prerequisites are satisfied.
The environment rejects such actions with an error message,
but neither terminates the rollout nor changes its state.
Consequently,
a rollout may repeatedly take invalid actions 
while leaving both its final utility and its modeled adaptation cost unchanged.
Therefore, 
we introduce \(P_{\mathrm{inv}}(\tau)\)
to provide a direct and dense learning signal
that favors clean and valid trajectories reaching the same outcome more efficiently.
We keep this penalty small and capped so that it shapes execution without overwhelming the utility signal or discouraging the use of TTT actions.
Specifically, \(P_{\mathrm{inv}}\) adds \(0.5\) for each rejected action, 
with its total capped at \(2.0\);
we set \(\lambda_{\mathrm{inv}}=0.15\) through all experiments unless otherwise specified.

Within each rollout group, 
we compute the mean-centered advantage following Dr.~GRPO~\citep{liu2025understanding}.
We omit group standard-deviation normalization to preserve the magnitude of reward differences,
and avoid implicitly reweighting queries by reward variance:
\begin{equation}
    A_t^{(g)}
    =
    R
    \left(
        \tau_t^{(g)}
    \right)
    -
    \frac{1}{G}
    \sum_{j=1}^{G}
    R
    \left(
        \tau_t^{(j)}
    \right).
    \label{eq:grpo_advantage}
\end{equation}
\vspace{-1em}

Using \(A_t^{(g)}\), we optimize the standard clipped GRPO objective over policy-generated reasoning and action tokens with a KL regularizer:
\begin{equation}
    \mathcal L_{\mathrm{RL}}(\theta)
    =
    -
    \mathbb E_{t,g,i}
    \left[
        \min
        \left(
            \rho_{t,g,i}(\theta)A_t^{(g)},
            \operatorname{clip}
            \left(
                \rho_{t,g,i}(\theta),
                1-\epsilon,
                1+\epsilon
            \right)
            A_t^{(g)}
        \right)
    \right]
    +
    \beta\mathcal L_{\mathrm{KL}}(\theta),
    \label{eq:outcome_rl_objective}
\end{equation}
where \(\rho_{t,g,i}\) is the probability ratio between the current and rollout policies for the \(i\)-th generated token,
and \(\mathcal L_{\mathrm{KL}}\) regularizes the policy toward the frozen base distribution.

Together,
the two training stages produce a policy
that can reliably decide whether and how to use TTT.
Guidance distillation first makes valid TTT trajectories executable and likely to be sampled, 
while outcome-driven reinforcement learning calibrates these decisions
using utility gains,
computation costs,
and invalid-action penalties.

\section{Experiments}

In this section, 
we ask whether Agentic-TTT can achieve self-improvement through learned agency over TTT skills;
whether its inference-time compute can be controlled;
and whether the learned policy transfers beyond the training distributions.

\paragraph{Data and evaluation.}
Our dataset covers eight domains
drawn from \(11\) benchmark families.
Guidance distillation uses \(2{,}072\) turn-level examples
from \(177\) trajectories,
and reinforcement learning uses a stream of \(187\) training inputs.
Evaluation follows a fixed sequential stream
containing \(176\) held-out scored queries,
starting with an empty skill library.
Library state is not shared across training or evaluation runs.
Overall utility is the mean binary score
assigned by task-specific automatic verifiers.
Appendix~\ref{sec:benchmark_evaluation_setup}
details the data sources,
splits,
stream construction,
and scoring rules.

\paragraph{Implementation.}
Unless otherwise specified,
we use Qwen2.5-7B-Instruct~\citep{hui2024qwen2}
with five TTT methods:
\textsc{PoT}~\citep{jiao2026pot},
\textsc{TTT-FS}~\citep{akyurek2025fsl},
\textsc{TTRL}~\citep{zuo2025ttrl},
\textsc{SEAL}~\citep{zweiger2025seal},
and \textsc{TLM}~\citep{hu2025tlm}.
The policy and TTT skills use separate LoRAs;
final-answer decoding disables the policy LoRA
and receives only the original query.
Both policy and answer decoding are greedy at evaluation.
For trained policies,
we report the mean and sample standard deviation
over three independent training runs,
evaluating each checkpoint once.
Harness and optimization details appear in
Appendices~\ref{sec:app:training}
and~\ref{sec:app:harness}.

\subsection{Main Results: Agentic-TTT Learns When and How to Apply TTT}

\begin{table}[t]
    \centering
    \caption{
        \textbf{Main results on Qwen2.5-7B.}
        Task rows report absolute utility for Direct Answer
        and mean changes relative to it for all other methods;
        Overall reports absolute utility.
        $\pm$ denotes the sample standard deviation over three training runs,
        and boldface marks the best non-oracle result.
    }
    \label{tab:main_results}

    \scriptsize
    \setlength{\tabcolsep}{1.5pt}
    \renewcommand{\arraystretch}{1.10}

    \begin{tabular}{
        l
        cccccc
        !{\vrule width 1.0pt}
        cc
        !{\vrule width 1.0pt}
        >{\columncolor[gray]{0.94}}c
    }
        \toprule

        &
        \multicolumn{1}{c}{Absolute score}
        &
        \multicolumn{8}{c}{
            Relative score ($\Delta$ vs.\ Direct Answer)
        } \\

        \cmidrule(lr){2-2}
        \cmidrule(lr){3-10}

        &
        &
        \multicolumn{5}{c}{Baselines}
        &
        \multicolumn{2}{
            !{\vrule width 1.0pt}c!{\vrule width 1.0pt}
        }{Our training}
        &
        \multicolumn{1}{
            >{\columncolor[gray]{0.94}}c
        }{Reference} \\

        \cmidrule(lr){3-7}
        \cmidrule(lr){8-9}
        \cmidrule(lr){10-10}

        \shortstack[l]{Evaluation\\task}
        & \shortstack{Direct\\Answer}
        & Prompted
        & \shortstack{Semantic\\Router}
        & \shortstack{Always\\TLM}
        & \shortstack{Always\\SEAL}
        & \shortstack{Always\\TTRL}
        & \shortstack{SFT\\only}
        & \shortstack{\textbf{SFT+RL}\\\textbf{(Ours)}}
        & Oracle \\

        \midrule

        Math
        & 0.526
        & $-0.018$
        & $-0.053$
        & $0.000$
        & $-0.368$
        & $+0.105$
        & $0.000$
        & $\mathbf{+0.175}$
        & $+0.281$ \\

        Code
        & 0.250
        & $-0.042$
        & $+0.042$
        & $0.000$
        & $-0.236$
        & $-0.042$
        & $+0.042$
        & $\mathbf{+0.167}$
        & $+0.403$ \\

        General Reasoning
        & 0.167
        & $-0.014$
        & $+0.138$
        & $0.000$
        & $-0.028$
        & $0.000$
        & $+0.083$
        & $\mathbf{+0.208}$
        & $+0.472$ \\

        Grid Puzzles
        & 0.000
        & $+0.024$
        & $+0.310$
        & $0.000$
        & $+0.048$
        & $0.000$
        & $+0.214$
        & $\mathbf{+0.381}$
        & $+0.357$ \\

        Logic Games
        & 0.367
        & $+0.011$
        & $+0.156$
        & $0.000$
        & $0.000$
        & $+0.233$
        & $+0.133$
        & $\mathbf{+0.367}$
        & $+0.300$ \\

        Medical QA
        & 0.400
        & $+0.089$
        & $+0.100$
        & $+0.022$
        & $+0.022$
        & $+0.167$
        & $+0.133$
        & $\mathbf{+0.200}$
        & $+0.333$ \\

        Passage QA
        & 0.040
        & $+0.027$
        & $+0.027$
        & $0.000$
        & $\mathbf{+0.427}$
        & $+0.040$
        & $+0.240$
        & $+0.347$
        & $+0.453$ \\

        Long-context QA
        & 0.100
        & $+0.033$
        & $0.000$
        & $\mathbf{+0.767}$
        & $+0.100$
        & $0.000$
        & $+0.100$
        & $+0.133$
        & $+0.767$ \\

        \midrule

        \textbf{Overall (abs.)}
        & 0.256
        & \shortstack{0.271\\[-1pt]{\scriptsize $\pm 0.003$}}
        & \shortstack{0.347\\[-1pt]{\scriptsize $\pm 0.020$}}
        & \shortstack{0.260\\[-1pt]{\scriptsize $\pm 0.003$}}
        & \shortstack{0.248\\[-1pt]{\scriptsize $\pm 0.029$}}
        & \shortstack{0.335\\[-1pt]{\scriptsize $\pm 0.038$}}
        & \shortstack{0.375\\[-1pt]{\scriptsize $\pm 0.010$}}
        & \shortstack{
            $\mathbf{0.510}$\\[-1pt]
            {\scriptsize $\pm 0.012$}
        }
        & \shortstack{0.650\\[-1pt]{\scriptsize $\pm 0.061$}} \\

        \bottomrule
    \end{tabular}
\end{table}

We compare all methods using the same backbone LLM and execution environment.
\textsc{Direct Answer} answers without any TTT,
while \textsc{Prompted} uses the same Agentic-TTT  prompt instruction but without any policy training.
\textsc{Always-TLM},
\textsc{Always-SEAL},
and \textsc{Always-TTRL} each use a fixed adaptation method.
\textsc{Semantic Router} selects a method from annotated training examples retrieved by query-embedding similarity.
Within our training pipeline,
\textsc{SFT-only} shows the result only after guidance distillation training,
whereas \textsc{SFT+RL} completes both training stages.
\textsc{Oracle} executes annotation-derived action sequences
for skill construction and reuse,
with methods selected through offline empirical profiling.
It serves as a non-deployable reference,
not a strict upper bound.

\begin{figure}[b]
    \centering
    \includegraphics[width=\textwidth]{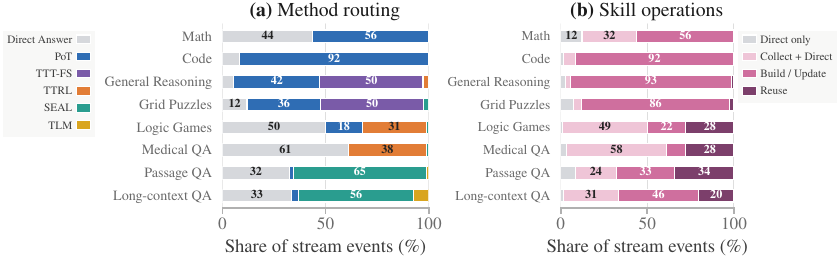}
    \caption{
        \textbf{Agentic-TTT learns task-dependent method routing and skill operations.}
        \textbf{(a)} Distribution of TTT methods for each response.
        \textbf{(b)} Distribution of skill operations for each input.
    }
    \label{fig:policy_behavior}
\end{figure}

Table~\ref{tab:main_results} shows that Agentic-TTT increases overall utility
from \(0.256\) to \(0.510\),
with positive gains in all eight domains.
Fixed-method benefits vary substantially across domains:
for example,
\textsc{Always-SEAL} improves Passage QA
but degrades mathematics and code generation.
Prompting and embedding-based routing provide smaller overall gains,
reaching \(0.271\) and \(0.347\), respectively.
Within our training pipeline,
guidance distillation raises utility to \(0.375\),
and outcome-driven RL further improves it to \(0.510\),
demonstrating additional benefit from optimizing measured trajectory outcomes.
Agentic-TTT also exceeds the annotation-derived \textsc{Oracle}
on Grid Puzzles and Logic Games,
showing that the learned policy can discover more effective adaptation decisions than the prescribed reference actions.

Figure~\ref{fig:policy_behavior} reveals how the learned policy manages TTT across tasks.
As shown in Figure~\ref{fig:policy_behavior}(a),
the policy does not follow a rigid routing strategy,
but instead selects from a diverse portfolio of TTT methods,
while retaining the option to answer with the frozen backbone.
Across three evaluation runs,
Agentic-TTT converts \(141\) incorrect direct answers into correct ones,
while reversing only \(7\) correct direct answers,
yielding a help-to-harm ratio of approximately \(20{:}1\).
Figure~\ref{fig:policy_behavior}(b)
further shows that the policy learns more than method routing alone.
Across the three runs,
it answers \(100\) queries using previously constructed skills;
these reuse decisions help on \(61\) queries
and harm only one relative to Direct Answer.
Loading these skills requires no additional adaptation,
allowing their construction cost to be amortized across subsequent queries.

\subsection{Training Agentic-TTT to Balance Utility and Compute}

To study whether Agentic-TTT can explicitly control its adaptation compute,
we train a family of policies with compute penalties
\(\lambda_{\mathrm{comp}} \in \{0, 0.1, 0.2, 0.4\}\).
For each TTT method \(m\),
we measure the FLOPs required for one successful adaptation,
and normalize them by the FLOPs of one direct answer,
yielding a method-specific normalized cost \(c_m\).
During guidance distillation,
we retain a TTT trajectory only when its empirical utility gain amortized over expected future reuse,
justifies its compute cost \(\lambda_{\mathrm{comp}}\log_{10}(1+c_m)\);
otherwise, we use a direct-answer trajectory.
During RL,
we subtract the same cost term from the sequence-level advantage.
The setting \(\lambda_{\mathrm{comp}}=0\) therefore serves as a compute-unconstrained,
utility-only reference under the same training pipeline.

\begin{figure}[t]
    \centering
    \includegraphics[width=\textwidth]
        {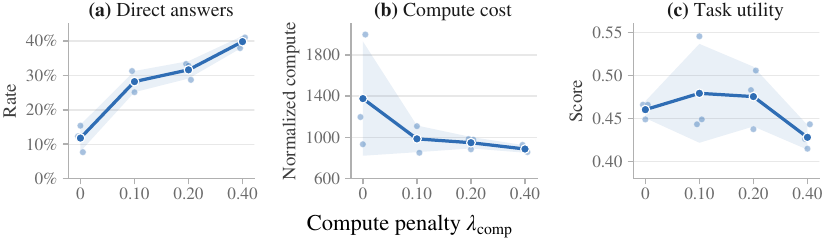}
    \caption{
        \textbf{Compute penalties steer the policy toward selective adaptation.}
        We vary the compute penalty~$\lambda_{\mathrm{comp}}$ and report
        (a)~the direct-answer rate,
        (b)~normalized adaptation compute,
        and (c)~task utility.
        Dots show \(3\) individual runs, lines and bands show their means and standard deviation.
    }
    \label{fig:compute-aware-behavior}
\end{figure}
\begin{figure}[b]
    \centering
    \includegraphics[width=\textwidth]
        {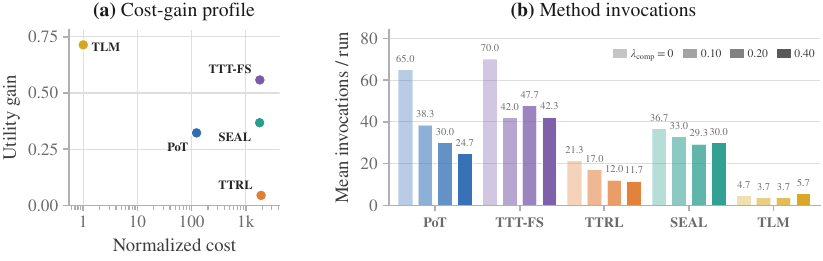}
    \caption{
        \textbf{Compute penalties selectively reallocate TTT method use.}
        \textbf{(a)}
        Normalized adaptation cost and mean training-set utility gain
        for each TTT method.
        \textbf{(b)}
        Average number of method invocations per evaluation run.
        Colors identify TTT methods,
        while darker shades indicate larger
        \(\lambda_{\mathrm{comp}}\).
    }
    \label{fig:compute-aware-methods}
\end{figure}

Figure~\ref{fig:compute-aware-behavior} shows a clear behavioral response to the compute penalty.
As \(\lambda_{\mathrm{comp}}\) increases from \(0\) to \(0.4\),
the mean direct-answer rate rises monotonically from \(11.8\%\) to \(39.8\%\),
while normalized adaptation compute decreases
from \(1375.5\) to \(886.9\),
a reduction of \(35.5\%\).
Interestingly, 
moderate penalties like \(\lambda_{\mathrm{comp}}=0.1\) and \(0.2\),
yield particularly favorable trade-offs:
they reduce compute by approximately \(30\%\) while achieving higher mean utility than \(\lambda_{\mathrm{comp}}=0\).
This suggests that without an explicit compute cost,
the policy may overuse TTT and become less selective about when adaptation is beneficial.

Figure~\ref{fig:compute-aware-methods} examines how this compute reduction is distributed across TTT methods.
As shown in Figure~\ref{fig:compute-aware-methods}(a),
different TTT methods have markedly different cost-gain profiles.
For example,
TTRL incurs a high adaptation cost but a marginal utility gain,
whereas TLM lies at the other extreme.
Importantly,
these utility gains are measured only on the training distributions matched to each method.
When a method is applied to unsuitable inputs,
its realized gain may be substantially smaller or even negative.
Figure~\ref{fig:compute-aware-methods}(b) shows that,
under stronger compute penalties,
the policy selectively reduces costly methods with limited expected benefit.
TTRL usage nearly halves,
decreasing from \(21.3\) to \(11.7\) invocations,
whereas TLM usage increases slightly.
This shift is consistent with the policy reallocating adaptation
toward less expensive methods.
These method-specific shifts show that Agentic-TTT learns a benefit-aware allocation of adaptation compute,
rather than merely suppressing TTT uniformly.

\subsection{Extrapolating TTT Agency to Unseen Data Sources}
\begin{figure}[h]
    \centering
    \includegraphics[width=\textwidth]{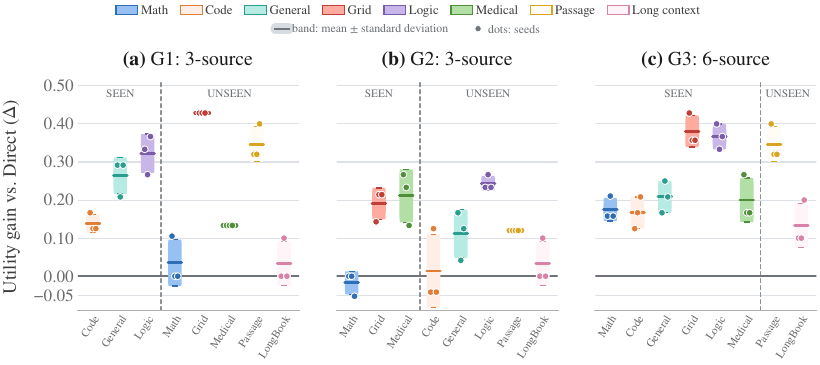}
    \caption{
        \textbf{Generalization beyond policy-training domains.}
        We report utility gains for policies trained on
        G1 (Code, General Reasoning, and Logic Games),
        G2 (Math, Grid Puzzles, and Medical QA),
        or their union G3.
        Passage and LongBook QA are unseen by all \(3\) groups.
        Bands and dots show the mean \(\pm\) std
        and individual results over three training runs.
    }
    \label{fig:training-data-extrapolation}
\end{figure}

Since test inputs may often fall outside the training distribution
and cannot be fully anticipated,
the learned TTT agency must generalize beyond data sources observed during training.
To evaluate this capability,
we construct three source-restricted training regimes.
G1 is trained only on Code,
General Reasoning,
and Logic Games,
whereas G2 is trained only on Math,
Grid Puzzles,
and Medical QA.
G3 is trained on the union of these six domains.
For each regime,
the excluded domains are absent from both guidance distillation
and reinforcement learning.
Passage QA and LongBook are additionally withheld from all three regimes,
providing a common set of jointly unseen domains.
We evaluate every trained policy on the same complete eight-domain suite
and report its utility gain relative to Direct Answer within each domain.

Figure~\ref{fig:training-data-extrapolation}
shows that the learned TTT agency is not confined to the data sources used for policy training.
For both G1 and G2,
the gains on held-out domains are broadly comparable
to those on domains seen during training.
In some cases,
transfer is particularly strong:
G1 achieves its largest improvement on Grid Puzzles,
despite never observing this domain during training.
Finally,
G3 performs best on Passage and LongBook QA,
which are unseen by all three policies,
suggesting that greater diversity in the training sources
supports stronger extrapolation to entirely new domains.
Overall,
Agentic-TTT learns transferable principles for selecting
and managing TTT skills,
rather than merely memorizing source-specific action patterns.

\subsection{Generalization Across Model Families and Scales}
\begin{table}[t]
    \centering

    \begin{minipage}[t]{0.37\linewidth}
        \vspace{0pt}
        \centering
        \textbf{(a) Overall utility}\\[3pt]

        \scriptsize
        \setlength{\tabcolsep}{2.5pt}
        \renewcommand{\arraystretch}{1.20}

        \begin{tabular}{@{}lrrr@{}}
            \toprule
            &
            \multicolumn{1}{c}{Absolute}
            &
            \multicolumn{2}{c}{Utility gain vs.\ Direct} \\
            \cmidrule(lr){2-2}
            \cmidrule(l){3-4}

            Base model
            & Direct
            & Prompted
            & \textbf{Ours} \\
            \midrule

            Llama-8B
                & .284
                & +.063
                & \textbf{+.136} \\

            GLM-9B
                & .341
                & \phantom{+}.000
                & \textbf{+.142} \\

            Qwen-7B
                & .256
                & +.015
                & \textbf{+.254} \\

            Qwen-14B
                & .432
                & $-$.136
                & \textbf{+.189} \\

            \bottomrule
        \end{tabular}
    \end{minipage}
    \hfill
    \begin{minipage}[t]{0.61\linewidth}
        \vspace{0pt}
        \centering
        \textbf{(b) Per-domain utility gain of our policy}\\[3pt]

        \scriptsize
        \setlength{\tabcolsep}{1.25pt}
        \renewcommand{\arraystretch}{1.20}

        \begin{tabular}{@{}lrrrrrrrr@{}}
            \toprule
            &
            \multicolumn{8}{c}{Utility gain vs.\ Direct} \\
            \cmidrule(l){2-9}

            Base model
            & Math
            & Code
            & Gen.
            & Grid
            & Logic
            & Med.
            & Pass.
            & Book \\
            \midrule

            Llama-8B
                & $-$.105
                & $-$.042
                & \phantom{+}.000
                & +.214
                & +.233
                & +.167
                & +.440
                & +.100 \\

            GLM-9B
                & +.211
                & +.042
                & +.292
                & +.286
                & $-$.033
                & \phantom{+}.000
                & +.480
                & $-$.200 \\

            Qwen-7B
                & +.175
                & +.167
                & +.208
                & +.381
                & +.367
                & +.200
                & +.347
                & +.133 \\

            Qwen-14B
                & +.158
                & +.222
                & +.250
                & +.405
                & $-$.044
                & $-$.078
                & +.653
                & +.067 \\

            \bottomrule
        \end{tabular}
    \end{minipage}

    \caption{
        \textbf{Generalization across model families and scales.}
        Direct reports the absolute utility of each backbone;
        Prompted and Ours report utility gains relative to Direct Answer.
        Panel (a) reports overall utility,
        while panel (b) reports the per-domain gains of our learned policy.
    }
    \label{tab:cross-model}
\end{table}

We evaluate Llama-3.1-8B and GLM-4-9B
to assess generality across model families,
and Qwen2.5-14B to assess gains with a stronger backbone.
Each policy is compared against baselines using the same backbone,
with all other settings identical to the main experiments.

\paragraph{Cross-model family.}
Agentic-TTT improves overall utility by \(0.254\) on Qwen-7B,
\(0.136\) on Llama-8B,
and \(0.142\) on GLM-9B,
substantially outperforming prompting alone.
This confirms that the learned TTT agency is not specific to Qwen2.5.
However,
the gains are model-dependent:
Llama degrades on mathematics and code,
while GLM degrades on Logic Games and Long-context QA.
These failures indicate that the policy does not always abstain from TTT when direct answering would be more reliable,
leaving room for better backbone-aware decision making.

\paragraph{Cross-model size.}
The gains also persist as the backbone becomes stronger.
Qwen-14B already achieves \(0.432\) utility through direct answering,
yet Agentic-TTT provides a further gain of \(0.189\),
raising its overall utility to \(0.621\).
Its improvement profile differs from that of Qwen-7B,
suggesting that stronger models retain different capability gaps.
However,
negative gains on Logic Games and Medical QA
highlight the need for better calibration
of adaptation decisions on stronger backbones.

\subsection{Ablation studies}

\subsubsection{Action-Token Supervision during SFT}

We compare \(\omega_a=1\) and \(4\),
and apply the same subsequent RL recipe to both checkpoints.
In Table~\ref{tab:action-supervision-weight},
invalid actions are generated action blocks that fail parsing or action-schema validation.
Increasing \(\omega_a\) from \(1\) to \(4\) improves final utility by \(0.046\) and reduces the invalid-action rate
from \(4.08\%\) to \(1.72\%\).
Thus,
upweighting the relatively sparse action tokens can increase action reliability and translate into better downstream utility.

\begin{table}[h]
    \centering

    \begin{minipage}[t]{0.485\textwidth}
        \vspace{0pt}
        \centering
        \caption{
            \textbf{SFT action-token supervision.}
        }
        \label{tab:action-supervision-weight}

        \small
        \setlength{\tabcolsep}{4pt}
        \renewcommand{\arraystretch}{1.08}

        \begin{tabular}{lcc}
            \toprule
            \shortstack[l]{\(\omega_a\)}
            & \shortstack{Utility $\uparrow$}
            & \shortstack{Invalid\\actions (\%) $\downarrow$} \\
            \midrule
            $1$
            & $0.464 \pm 0.017$
            & $4.08 \pm 1.30$ \\
            $4$ (ours)
            & $\mathbf{0.510 \pm 0.012}$
            & $\mathbf{1.72 \pm 0.55}$ \\
            \bottomrule
        \end{tabular}

    \end{minipage}
    \hfill
    \begin{minipage}[t]{0.485\textwidth}
        \vspace{0pt}
        \centering
        \caption{
            \textbf{RL invalid-action penalty.}
        }
        \label{tab:invalid-action-penalty}

        \small
        \setlength{\tabcolsep}{4pt}
        \renewcommand{\arraystretch}{1.08}

        \begin{tabular}{lcc}
            \toprule
            \shortstack[l]{\(\lambda_{\mathrm{inv}}\)}
            & \shortstack{Utility $\uparrow$}
            & \shortstack{Budget\\exhausted (\%)}$\downarrow$ \\
            \midrule
            $0$
            & $0.467 \pm 0.027$
            & $21.20 \pm 0.59$ \\
            $0.15$ (ours)
            & $\mathbf{0.510 \pm 0.012}$
            & $\mathbf{13.33 \pm 2.07}$ \\
            \bottomrule
        \end{tabular}
    \end{minipage}
\end{table}

\subsubsection{Invalid Action Penalty in RL}

We compare RL training with and without the invalid-action penalty.
The budget-exhaustion rate in Table~\ref{tab:invalid-action-penalty} is the fraction of trajectories that consume their entire 6 action budget.
Setting \(\lambda_{\mathrm{inv}}=0.15\) improves utility by \(0.043\) and reduces budget exhaustion by \(7.87\) percentage points.
The penalty therefore helps the policy avoid unproductive action sequences and preserve its finite action budget for completing the task.

\section{Conclusion}
Our results suggest that enabling neural models to grow after deployment requires more than the ability to update their parameters; 
it requires agency over those updates.
Agentic-TTT provides this agency through a learned test-time policy that manages heterogeneous TTT procedures and reusable parameter-space skills across a stream of queries.
Experiments show that no fixed TTT method is broadly effective, whereas Agentic-TTT nearly doubles utility over direct inference, 
balances utility against compute, 
and transfers to domains unseen during training.
Taken together, this work reframes TTT from a fixed adaptation procedure into a learned decision process, 
pointing toward models that can decide how to learn from their own deployment experience.
\clearpage

\newpage
\bibliography{iclr2027_conference}

@inproceedings{zuo2025ttrl,
  title={TTRL: Test-Time Reinforcement Learning},
  author={Zuo, Yuxin and Zhang, Kaiyan and Sheng, Li and Qu, Shang and Cui, Ganqu and Zhu, Xuekai and Li, Haozhan and Zhang, Yuchen and Long, Xinwei and Hua, Ermo and others},
  booktitle={The Thirty-ninth Annual Conference on Neural Information Processing Systems},
  year={2025}
}

@inproceedings{zweiger2025seal,
  title={Self-Adapting Language Models},
  author={Zweiger, Adam and Pari, Jyothish and Guo, Han and Kim, Yoon and Agrawal, Pulkit},
  booktitle={The Thirty-ninth Annual Conference on Neural Information Processing Systems},
  year={2025}
}

@inproceedings{hu2025tlm,
  title={Test-Time Learning for Large Language Models},
  author={Hu, Jinwu and Zhang, Zitian and Chen, Guohao and Wen, Xutao and Shuai, Chao and Luo, Wei and Xiao, Bin and Li, Yuanqing and Tan, Mingkui},
  booktitle={International Conference on Machine Learning},
  pages={24823--24849},
  year={2025},
  organization={PMLR}
}

@inproceedings{akyurek2025fsl,
  title={The Surprising Effectiveness of Test-Time Training for Few-Shot Learning},
  author={Aky{\"u}rek, Ekin and Damani, Mehul and Zweiger, Adam and Qiu, Linlu and Guo, Han and Pari, Jyothish and Kim, Yoon and Andreas, Jacob},
  booktitle={International Conference on Machine Learning},
  pages={942--963},
  year={2025},
  organization={PMLR}
}

@article{yuksekgonul2026discover,
  title={Learning to discover at test time},
  author={Yuksekgonul, Mert and Koceja, Daniel and Li, Xinhao and Bianchi, Federico and McCaleb, Jed and Wang, Xiaolong and Kautz, Jan and Choi, Yejin and Zou, James and Guestrin, Carlos and others},
  journal={arXiv preprint arXiv:2601.16175},
  year={2026}
}

@article{hubert2025olympiad,
  title={Olympiad-level formal mathematical reasoning with reinforcement learning},
  author={Hubert, Thomas and Mehta, Rishi and Sartran, Laurent and Horv{\'a}th, Mikl{\'o}s Z and {\v{Z}}u{\v{z}}i{\'c}, Goran and Wieser, Eric and Huang, Aja and Schrittwieser, Julian and Schroecker, Yannick and Masoom, Hussain and others},
  journal={Nature},
  pages={1--3},
  year={2025},
  publisher={Nature Publishing Group UK London}
}

@inproceedings{moradi2025continuous,
title={Continuous Self-Improvement of Large Language Models by Test-time Training with Verifier-Driven Sample Selection},
author={Mohammad Mahdi Moradi and Hossam Amer and Sudhir Mudur and Weiwei Zhang and Yang Liu and Walid Ahmed},
booktitle={AI That Keeps Up: NeurIPS 2025 Workshop on Continual and Compatible Foundation Model Updates},
year={2025},
url={https://openreview.net/forum?id=6ahliSpvQ0}
}

@article{hubotter2025learning,
  title={Learning on the job: Test-time curricula for targeted reinforcement learning},
  author={H{\"u}botter, Jonas and Diaz-Bone, Leander and Hakimi, Ido and Krause, Andreas and Hardt, Moritz},
  journal={arXiv preprint arXiv:2510.04786},
  year={2025}
}

@article{liu2025ettrl,
  title={Ettrl: Balancing exploration and exploitation in llm test-time reinforcement learning via entropy mechanism},
  author={Liu, Jia and He, ChangYi and Lin, YingQiao and Yang, MingMin and Shen, FeiYang and Liu, ShaoGuo},
  journal={arXiv preprint arXiv:2508.11356},
  year={2025}
}

@inproceedings{sun2025learning,
  title={Learning to (Learn at Test Time): RNNs with Expressive Hidden States},
  author={Sun, Yu and Li, Xinhao and Dalal, Karan and Xu, Jiarui and Vikram, Arjun and Zhang, Genghan and Dubois, Yann and Chen, Xinlei and Wang, Xiaolong and Koyejo, Sanmi and others},
  booktitle={International Conference on Machine Learning},
  pages={57503--57522},
  year={2025},
  organization={PMLR}
}

@article{zhang2025test,
  title={Test-time training done right},
  author={Zhang, Tianyuan and Bi, Sai and Hong, Yicong and Zhang, Kai and Luan, Fujun and Yang, Songlin and Sunkavalli, Kalyan and Freeman, William T and Tan, Hao},
  journal={arXiv preprint arXiv:2505.23884},
  year={2025}
}

@article{tandon2025e2e,
  title={End-to-end test-time training for long context},
  author={Tandon, Arnuv and Dalal, Karan and Li, Xinhao and Koceja, Daniel and R{\o}d, Marcel and Buchanan, Sam and Wang, Xiaolong and Leskovec, Jure and Koyejo, Sanmi and Hashimoto, Tatsunori and others},
  journal={arXiv preprint arXiv:2512.23675},
  year={2025}
}

@inproceedings{feng2026inplace,
    title={In-Place Test-Time Training},
    author={Guhao Feng and Shengjie Luo and Kai Hua and Ge Zhang and Wenhao Huang and Di He and Tianle Cai},
    booktitle={The Fourteenth International Conference on Learning Representations},
    year={2026},
    url={https://openreview.net/forum?id=dTWfCLSoyl}
}

@article{shekar2025adaptive,
  title={Adaptive Minds: Empowering Agents with LoRA-as-Tools},
  author={Shekar, Pavan C and Krishnan, Ashwanth},
  journal={arXiv preprint arXiv:2510.15416},
  year={2025}
}

@inproceedings{huang2024lorahub,
  title={LoraHub: Efficient Cross-Task Generalization via Dynamic LoRA Composition},
  author={Huang, Chengsong and Liu, Qian and Lin, Bill Yuchen and Pang, Tianyu and Du, Chao and Lin, Min},
  booktitle={First Conference on Language Modeling},
  year={2024}
}

@inproceedings{ostapenko2024towards,
  title={Towards Modular LLMs by Building and Reusing a Library of LoRAs},
  author={Ostapenko, Oleksiy and Su, Zhan and Ponti, Edoardo and Charlin, Laurent and Le Roux, Nicolas and Caccia, Lucas and Sordoni, Alessandro},
  booktitle={International Conference on Machine Learning},
  pages={38885--38904},
  year={2024},
  organization={PMLR}
}

@inproceedings{huang2024mixture,
  title={Mixture of lora experts},
  author={Huang, Shaohan and Wei, Furu and others},
  booktitle={International Conference on Learning Representations},
  volume={2024},
  pages={47302--47318},
  year={2024}
}

@article{li2024mixlora,
  title={Mixlora: Enhancing large language models fine-tuning with lora-based mixture of experts},
  author={Li, Dengchun and Ma, Yingzi and Wang, Naizheng and Ye, Zhengmao and Cheng, Zhiyuan and Tang, Yinghao and Zhang, Yan and Duan, Lei and Zuo, Jie and Yang, Cal and others},
  journal={arXiv preprint arXiv:2404.15159},
  year={2024}
}

@inproceedings{bertolissi2025local,
  title={Local Mixtures of Experts: Essentially Free Test-Time Training via Model Merging},
  author={Bertolissi, Ryo and H{\"u}botter, Jonas and Hakimi, Ido and Krause, Andreas},
  booktitle={Second Conference on Language Modeling},
  year={2025},
  organization={OpenReview}
}

@inproceedings{charakorn2025text,
  title={Text-to-LoRA: Instant Transformer Adaption},
  author={Charakorn, Rujikorn and Cetin, Edoardo and Tang, Yujin and Lange, Robert Tjarko},
  booktitle={International Conference on Machine Learning},
  pages={7485--7514},
  year={2025},
  organization={PMLR}
}

@inproceedings{wei2025online,
  title={Online-lora: Task-free online continual learning via low rank adaptation},
  author={Wei, Xiwen and Li, Guihong and Marculescu, Radu},
  booktitle={2025 IEEE/CVF Winter Conference on Applications of Computer Vision (WACV)},
  pages={6634--6645},
  year={2025},
  organization={IEEE}
}

@article{acikgoz2025self,
  title={Self-improving llm agents at test-time},
  author={Acikgoz, Emre Can and Qian, Cheng and Ji, Heng and Hakkani-T{\"u}r, Dilek and Tur, Gokhan},
  journal={arXiv preprint arXiv:2510.07841},
  year={2025}
}

@article{wei2025lightweight,
  title={A Lightweight Framework for Trigger-Guided LoRA-Based Self-Adaptation in LLMs},
  author={Wei, Jiacheng and Wu, Faguo and Zhang, Xiao},
  journal={arXiv preprint arXiv:2509.05385},
  year={2025}
}

@article{rusu2016progressive,
  title={Progressive neural networks},
  author={Rusu, Andrei A and Rabinowitz, Neil C and Desjardins, Guillaume and Soyer, Hubert and Kirkpatrick, James and Kavukcuoglu, Koray and Pascanu, Razvan and Hadsell, Raia},
  journal={arXiv preprint arXiv:1606.04671},
  year={2016}
}

@inproceedings{yoon2018lifelong,
  title={Lifelong learning with dynamically expandable networks},
  author={Yoon, Jaehong and Yang, Eunho and Lee, Jeongtae and Hwang, Sung Ju},
  booktitle={6th International Conference on Learning Representations, ICLR 2018},
  year={2018}
}

@article{ostapenko2021continual,
  title={Continual learning via local module composition},
  author={Ostapenko, Oleksiy and Rodriguez, Pau and Caccia, Massimo and Charlin, Laurent},
  journal={Advances in Neural Information Processing Systems},
  volume={34},
  pages={30298--30312},
  year={2021}
}

@inproceedings{wang2025self,
  title={Self-expansion of pre-trained models with mixture of adapters for continual learning},
  author={Wang, Huiyi and Lu, Haodong and Yao, Lina and Gong, Dong},
  booktitle={Proceedings of the Computer Vision and Pattern Recognition Conference},
  pages={10087--10098},
  year={2025}
}

@inproceedings{ellis2021dreamcoder,
  title={Dreamcoder: Bootstrapping inductive program synthesis with wake-sleep library learning},
  author={Ellis, Kevin and Wong, Catherine and Nye, Maxwell and Sabl{\'e}-Meyer, Mathias and Morales, Lucas and Hewitt, Luke and Cary, Luc and Solar-Lezama, Armando and Tenenbaum, Joshua B},
  booktitle={Proceedings of the 42nd acm sigplan international conference on programming language design and implementation},
  pages={835--850},
  year={2021}
}

@inproceedings{grand2024lilo,
  title={Lilo: Learning interpretable libraries by compressing and documenting code},
  author={Grand, Gabriel and Wong, Lio and Bowers, Maddy and Olausson, Theo X and Liu, Muxin and Tenenbaum, Joshua B and Andreas, Jacob},
  booktitle={International Conference on Learning Representations},
  volume={2024},
  pages={30399--30446},
  year={2024}
}

@article{wang2024voyager,
title={Voyager: An Open-Ended Embodied Agent with Large Language Models},
author={Guanzhi Wang and Yuqi Xie and Yunfan Jiang and Ajay Mandlekar and Chaowei Xiao and Yuke Zhu and Linxi Fan and Anima Anandkumar},
journal={Transactions on Machine Learning Research},
issn={2835-8856},
year={2024},
url={https://openreview.net/forum?id=ehfRiF0R3a},
note={}
}

@article{zheng2025skillweaver,
  title={Skillweaver: Web agents can self-improve by discovering and honing skills},
  author={Zheng, Boyuan and Fatemi, Michael Y and Jin, Xiaolong and Wang, Zora Zhiruo and Gandhi, Apurva and Song, Yueqi and Gu, Yu and Srinivasa, Jayanth and Liu, Gaowen and Neubig, Graham and others},
  journal={arXiv preprint arXiv:2504.07079},
  year={2025}
}

@inproceedings{wan2024lotus,
  title={Lotus: Continual imitation learning for robot manipulation through unsupervised skill discovery},
  author={Wan, Weikang and Zhu, Yifeng and Shah, Rutav and Zhu, Yuke},
  booktitle={2024 IEEE International Conference on Robotics and Automation (ICRA)},
  pages={537--544},
  year={2024},
  organization={IEEE}
}

@inproceedings{tziafas2024lifelong,
  title={Lifelong robot library learning: Bootstrapping composable and generalizable skills for embodied control with language models},
  author={Tziafas, Georgios and Kasaei, Hamidreza},
  booktitle={2024 IEEE International Conference on Robotics and Automation (ICRA)},
  pages={515--522},
  year={2024},
  organization={IEEE}
}

@techreport{vonneumann1945first,
  author    = {Von Neumann, John},
  title     = {First Draft of a Report on the {EDVAC}},
  institution = {Moore School of Electrical Engineering, University of Pennsylvania},
  year      = {1945}
}

@incollection{turing1947lecture,
  author    = {Turing, Alan M.},
  title     = {Lecture to the {London Mathematical Society}, 20 {February} 1947},
  booktitle = {A.~M.~Turing's {ACE} Report of 1946 and Other Papers},
  editor    = {Carpenter, B. E. and Doran, R. W.},
  publisher = {MIT Press},
  year      = {1986}
}

@inproceedings{niu2022efficient,
  title={Efficient test-time model adaptation without forgetting},
  author={Niu, Shuaicheng and Wu, Jiaxiang and Zhang, Yifan and Chen, Yaofo and Zheng, Shijian and Zhao, Peilin and Tan, Mingkui},
  booktitle={International conference on machine learning},
  pages={16888--16905},
  year={2022},
  organization={PMLR}
}

@inproceedings{niu2023towards,
  title={Towards Stable Test-time Adaptation in Dynamic Wild World},
  author={Niu, Shuaicheng and Wu, Jiaxiang and Zhang, Yifan and Wen, Zhiquan and Chen, Yaofo and Zhao, Peilin and Tan, Mingkui},
  booktitle={The Eleventh International Conference on Learning Representations},
  year={2023}
}

@article{gong2023sotta,
  title={Sotta: Robust test-time adaptation on noisy data streams},
  author={Gong, Taesik and Kim, Yewon and Lee, Taeckyung and Chottananurak, Sorn and Lee, Sung-Ju},
  journal={Advances in Neural Information Processing Systems},
  volume={36},
  pages={14070--14093},
  year={2023}
}

@article{hu2022lora,
  title={Lora: Low-rank adaptation of large language models.},
  author={Hu, Edward J and Shen, Yelong and Wallis, Phillip and Allen-Zhu, Zeyuan and Li, Yuanzhi and Wang, Shean and Wang, Liang and Chen, Weizhu and others},
  journal={Iclr},
  volume={1},
  number={2},
  pages={3},
  year={2022}
}

@inproceedings{pfeiffer2021adapterfusion,
  title={Adapterfusion: Non-destructive task composition for transfer learning},
  author={Pfeiffer, Jonas and Kamath, Aishwarya and R{\"u}ckl{\'e}, Andreas and Cho, Kyunghyun and Gurevych, Iryna},
  booktitle={Proceedings of the 16th conference of the European chapter of the association for computational linguistics: main volume},
  pages={487--503},
  year={2021}
}

@article{dhasade2026effective,
  title={Effective LoRA Adapter Routing using Task Representations},
  author={Dhasade, Akash and Kermarrec, Anne-Marie and Pavlovic, Igor and Petrescu, Diana and Pires, Rafael and Randl, Mathis and de Vos, Martijn},
  journal={arXiv preprint arXiv:2601.21795},
  year={2026}
}

@inproceedings{wang2025inducing,
  title={Inducing Programmatic Skills for Agentic Tasks},
  author={Wang, Zora Zhiruo and Gandhi, Apurva and Neubig, Graham and Fried, Daniel},
  booktitle={Second Conference on Language Modeling},
  year={2025}
}

@inproceedings{wang2026reinforcement,
  title={Reinforcement learning for self-improving agent with skill library},
  author={Wang, Jiongxiao and Yan, Qiaojing and Wang, Yawei and Tian, Yijun and Mishra, Soumya Smruti and Xu, Zhichao and Gandhi, Megha and Xu, Panpan and Cheong, Lin Lee},
  booktitle={Proceedings of the 64th Annual Meeting of the Association for Computational Linguistics (Volume 1: Long Papers)},
  pages={1529--1550},
  year={2026}
}

@article{shinn2023reflexion,
  title={Reflexion: Language agents with verbal reinforcement learning},
  author={Shinn, Noah and Cassano, Federico and Gopinath, Ashwin and Narasimhan, Karthik and Yao, Shunyu},
  journal={Advances in neural information processing systems},
  volume={36},
  pages={8634--8652},
  year={2023}
}

@inproceedings{zhao2024expel,
  title={Expel: Llm agents are experiential learners},
  author={Zhao, Andrew and Huang, Daniel and Xu, Quentin and Lin, Matthieu and Liu, Yong-Jin and Huang, Gao},
  booktitle={Proceedings of the AAAI Conference on Artificial Intelligence},
  volume={38},
  number={17},
  pages={19632--19642},
  year={2024}
}

@inproceedings{fernando2024promptbreeder,
  title={Promptbreeder: self-referential self-improvement via prompt evolution},
  author={Fernando, Chrisantha and Banarse, Dylan and Michalewski, Henryk and Osindero, Simon and Rockt{\"a}schel, Tim},
  booktitle={Proceedings of the 41st International Conference on Machine Learning},
  pages={13481--13544},
  year={2024}
}

@inproceedings{zhang2025aflow,
  title={Aflow: Automating agentic workflow generation},
  author={Zhang, Jiayi and Xiang, Jinyu and Yu, Zhaoyang and Teng, Fengwei and Chen, Xionghui and Chen, Jiaqi and Zhuge, Mingchen and Cheng, Xin and Hong, Sirui and Wang, Jinlin and others},
  booktitle={International Conference on Learning Representations},
  volume={2025},
  pages={34040--34077},
  year={2025}
}

@inproceedings{hu2025agentgen,
  title={Agentgen: Enhancing planning abilities for large language model based agent via environment and task generation},
  author={Hu, Mengkang and Zhao, Pu and Xu, Can and Sun, Qingfeng and Lou, Jian-Guang and Lin, Qingwei and Luo, Ping and Rajmohan, Saravan},
  booktitle={Proceedings of the 31st ACM SIGKDD Conference on Knowledge Discovery and Data Mining V. 1},
  pages={496--507},
  year={2025}
}

@article{zhao2026absolute,
  title={Absolute zero: Reinforced self-play reasoning with zero data},
  author={Zhao, Andrew and Wu, Yiran and Wu, Tong and Xu, Quentin and Yue, Yang and Lin, Matthieu and Wang, Shenzhi and Wu, Qingyun and Zheng, Zilong and Huang, Gao},
  journal={Advances in Neural Information Processing Systems},
  volume={38},
  pages={105816--105879},
  year={2026}
}

@inproceedings{liu2025understanding,
  title={Understanding R1-Zero-Like Training: A Critical Perspective},
  author={Liu, Zichen and Chen, Changyu and Li, Wenjun and Qi, Penghui and Pang, Tianyu and Du, Chao and Lee, Wee Sun and Lin, Min},
  booktitle={Second Conference on Language Modeling},
  year={2025}
}

@article{hui2024qwen2,
  title={Qwen2. 5-coder technical report},
  author={Hui, Binyuan and Yang, Jian and Cui, Zeyu and Yang, Jiaxi and Liu, Dayiheng and Zhang, Lei and Liu, Tianyu and Zhang, Jiajun and Yu, Bowen and Lu, Keming and others},
  journal={arXiv preprint arXiv:2409.12186},
  year={2024}
}

@article{jiao2026pot,
  title={Policy of Thoughts: Scaling LLM Reasoning via Test-time Policy Evolution},
  author={Jiao, Zhengbo and Xian, Hongyu and Wang, Qinglong and Ma, Yunpu and Wang, Zhebo and Zhang, Zifan and Kong, Dezhang and Han, Meng},
  journal={arXiv preprint arXiv:2601.20379},
  year={2026}
}

@article{duan2026last,
  title={The Last AI Built by Humans: Toward Genuine Recursive Self-Improvement},
  author={Duan, Yi and Liu, Ying and Tang, Zirui and Chen, Haodong and Zhou, Jun and Liu, Yumou and Xu, Bangrui and Wu, Yukai and Chen, Sidi and Zhou, Yuhan and others},
  journal={arXiv preprint arXiv:2609.11873},
  year={2026}
}

@inproceedings{lightman2024let,
  title={Let's verify step by step},
  author={Lightman, Hunter and Kosaraju, Vineet and Burda, Yuri and Edwards, Harrison and Baker, Bowen and Lee, Teddy and Leike, Jan and Schulman, John and Sutskever, Ilya and Cobbe, Karl},
  booktitle={International Conference on Learning Representations},
  volume={2024},
  pages={39578--39601},
  year={2024}
}

@misc{aimo2024amc,
  author       = {{AI-MO}},
  title        = {{aimo-validation-amc}},
  year         = {2024},
  howpublished = {Hugging Face dataset},
  url          = {https://huggingface.co/datasets/AI-MO/aimo-validation-amc}
}

@misc{maxwelljia2024aime,
  author       = {{Maxwell-Jia}},
  title        = {{AIME 2024 Dataset}},
  year         = {2024},
  howpublished = {Hugging Face dataset},
  url          = {https://huggingface.co/datasets/Maxwell-Jia/AIME_2024}
}

@inproceedings{jain2025livecodebench,
  title={Livecodebench: Holistic and contamination free evaluation of large language models for code},
  author={Jain, Naman and Gu, Alex and Li, Wen-Ding and Yan, Fanjia and Zhang, Tianjun and Wang, Sida and Solar-Lezama, Armando and Sen, Koushik and Stoica, Ion},
  booktitle={International Conference on Learning Representations},
  volume={2025},
  pages={58791--58831},
  year={2025}
}

@article{austin2021program,
  title={Program synthesis with large language models},
  author={Austin, Jacob and Odena, Augustus and Nye, Maxwell and Bosma, Maarten and Michalewski, Henryk and Dohan, David and Jiang, Ellen and Cai, Carrie and Terry, Michael and Le, Quoc and others},
  journal={arXiv preprint arXiv:2108.07732},
  year={2021}
}

@inproceedings{suzgun2023challenging,
  title={Challenging big-bench tasks and whether chain-of-thought can solve them},
  author={Suzgun, Mirac and Scales, Nathan and Sch{\"a}rli, Nathanael and Gehrmann, Sebastian and Tay, Yi and Chung, Hyung Won and Chowdhery, Aakanksha and Le, Quoc and Chi, Ed H and Zhou, Denny and others},
  booktitle={Findings of the Association for Computational Linguistics: ACL 2023},
  pages={13003--13051},
  year={2023}
}

@article{chollet2019measure,
  title={On the measure of intelligence},
  author={Chollet, Fran{\c{c}}ois},
  journal={arXiv preprint arXiv:1911.01547},
  year={2019}
}

@inproceedings{zhong2024agieval,
  title={Agieval: A human-centric benchmark for evaluating foundation models},
  author={Zhong, Wanjun and Cui, Ruixiang and Guo, Yiduo and Liang, Yaobo and Lu, Shuai and Wang, Yanlin and Saied, Amin and Chen, Weizhu and Duan, Nan},
  booktitle={Findings of the association for computational linguistics: NAACL 2024},
  pages={2299--2314},
  year={2024}
}

@inproceedings{pal2022medmcqa,
  title={Medmcqa: A large-scale multi-subject multi-choice dataset for medical domain question answering},
  author={Pal, Ankit and Umapathi, Logesh Kumar and Sankarasubbu, Malaikannan},
  booktitle={Conference on health, inference, and learning},
  pages={248--260},
  year={2022},
  organization={PMLR}
}

@inproceedings{rajpurkar2016squad,
  title={Squad: 100,000+ questions for machine comprehension of text},
  author={Rajpurkar, Pranav and Zhang, Jian and Lopyrev, Konstantin and Liang, Percy},
  booktitle={Proceedings of the 2016 conference on empirical methods in natural language processing},
  pages={2383--2392},
  year={2016}
}

@inproceedings{zhang2024bench,
  title={{$\infty$Bench}: Extending long context evaluation beyond 100K tokens},
  author={Zhang, Xinrong and Chen, Yingfa and Hu, Shengding and Xu, Zihang and Chen, Junhao and Hao, Moo and Han, Xu and Thai, Zhen and Wang, Shuo and Liu, Zhiyuan and others},
  booktitle={Proceedings of the 62nd Annual Meeting of the Association for Computational Linguistics (Volume 1: Long Papers)},
  pages={15262--15277},
  year={2024}
}
\bibliographystyle{iclr2027_conference}

\newpage
\appendix
\section*{Appendix}
The appendix is organized as follows.
Section~\ref{sec:related_work} reviews related work.
Section~\ref{sec:benchmark_evaluation_setup}
details benchmark construction.
Section~\ref{sec:app:training} provides detailed training implementations and configurations.
Section~\ref{sec:app:harness}
describes the execution harness and prompting configuration.
Section~\ref{sec:app:limitations}
discusses the limitations of this work
and outlines future research directions.

\section{Related Work}
\label{sec:related_work}
\subsection{Test-time Training for Language Models}

Test-time training(TTT) methods adapt LLMs during inference using signals derived from test inputs.
These signals include majority-vote pseudo-labels for reinforcement learning in TTRL~\citep{zuo2025ttrl}, 
model-generated self-edits in SEAL~\citep{zweiger2025seal}, 
input perplexity in TLM~\citep{hu2025tlm}, 
leave-one-out reconstruction of few-shot demonstrations~\citep{akyurek2025fsl} and more. 
Per-problem iterative search and adaptation in TTT-Discover~\citep{yuksekgonul2026discover} and 
auxiliary RL training in AlphaProof~\citep{hubert2025olympiad} 
have provided super-human level open scientific and mathematical problem solving. 
Related work further improves task-level adaptation through verifier-guided sample selection,
targeted test-time curricula, and exploration-aware reinforcement learning~\citep{moradi2025continuous,hubotter2025learning,liu2025ettrl}.

While these methods show strong empirical gains within a pre-specified task or evaluation settings like IMO competition~\citep{hubert2025olympiad},
in an open-world deployment, the assumption typically fails: 
live workloads contain free-form, unpredictable and highly diverse queries which cannot be assumed in advance.
Previous work have proposed selective test-time adaptation which filters uninformative samples within a fixed domain (primarily in computer vision)~\citep{niu2022efficient,niu2023towards,gong2023sotta},
In contrast, heterogeneous LLM workloads require higher-level autonomy: 
whether to enhance the model with TTT or just answer with the frozen base model, 
which TTT method to invoke, 
and whether an existing adaptation already suffices.
To our knowledge, no prior work on LLM test-time training has jointly addressed these gap.

A separate line of TTT uses test-time updated fast weights as internal memory for long-context modeling,
often integrating this mechanism into the model architecture or pretraining scheme~\citep{sun2025learning,zhang2025test,feng2026inplace,tandon2025e2e}.
This complementary line falls outside our scope:
fast-weight TTT designs the model's memory substrate, 
whereas Agentic-TTT governs the deployment-time acquisition and reuse of capabilities in post-trained LLMs.

\subsection{Modular Adapters and Self-Expanding Networks}

Modular adapters store specialized capabilities in parameter modules while keeping the base model frozen~\citep{hu2022lora, pfeiffer2021adapterfusion}.
Prior work builds libraries of such adapters and selects, routes, or composes them for different tasks~\citep{huang2024lorahub, ostapenko2024towards, shekar2025adaptive, bertolissi2025local, dhasade2026effective}.
Related approaches organize adapters as learned mixtures or generate them from task descriptions~\citep{huang2024mixture,li2024mixlora,charakorn2025text}.
These studies make model capabilities modular and reusable, but generally assume that the adapter pool or its generation mechanism is established before deployment.

Self-expanding networks tackle new tasks or distribution shifts by adding specialized modules rather than repeatedly overwriting a shared parameter state~\citep{rusu2016progressive, yoon2018lifelong,  ostapenko2021continual, wang2025self}.
Recent online and test-time adaptation methods bring this expansion into deployment~\citep{moradi2025continuous, wei2025online, acikgoz2025self}.
However, module creation is typically governed by predefined task boundaries or fixed adaptation rules; 
for example, SAGE~\citep{wei2025lightweight} triggers LoRA construction when reasoning failures are detected. 
Agentic-TTT instead turns test-time training into an agentic capability-management problem: 
the LLM policy not only selects among existing capabilities, but also decides when and how to expand a persistent skill library, 
thereby reshaping the capability space available to its future decisions.

\subsection{Self-Improving Agents}

Self-improving agents turn experience into reusable skills that can be stored, retrieved and applied across tasks.
These skills take diverse forms, including program and code libraries~\citep{ellis2021dreamcoder,grand2024lilo}, 
executable or text-described procedures~\citep{wang2024voyager,zheng2025skillweaver}, and reusable robot policies~\citep{wan2024lotus,tziafas2024lifelong}.
Recent systems go beyond accumulating skills to induce and verify them online, or to train agents to generate and use a growing skill library~\citep{wang2025inducing, wang2026reinforcement}.
Agentic-TTT extends this paradigm to parameter-space skills:
heterogeneous TTT methods create isolated LoRA adaptations, while a learned LLM policy governs their construction and reuse over a frozen base model,
supporting self-improvement during deployment~\citep{duan2026last}.

\section{Benchmark and Data Construction}
\label{sec:benchmark_evaluation_setup}

\subsection{Dataset Composition and Stream Construction}

Our main experiments cover eight domains drawn from eleven
benchmark families.
The dataset contains 229 training queries,
45 development queries,
and 176 test queries.
The training split provides the pool
from which subsets are drawn for guidance distillation
and reinforcement learning.
Guidance distillation uses 2,072 supervised turns collected from rollouts of training queries.
We use the development split
for validation and hyperparameter selection during policy training.
Development queries are excluded
from SFT and RL updates for the main policy.
LSAT-AR and MedMCQA provide sequential workloads for TTRL,
where the policy can accumulate related queries,
construct an adaptation,
and reuse it on subsequent queries.
For each of these two domains,
we construct one 30-query sequence for training
and another for testing.
Each sequence contains 16 initial queries
for collecting adaptation data,
followed by 14 queries for assessing skill reuse.
For MedMCQA,
the training and test sequences are drawn from
Pathology and Dental, respectively.

\begin{table}[h]
\centering
\small
\setlength{\tabcolsep}{4pt}
\caption{
Dataset composition across the eight domains.
Counts refer to queries in each split,
excluding document presentation events.
}
\label{tab:dataset_composition}
\begin{tabular}{@{}llrrr@{}}
\toprule
Domain & Dataset & Train queries & Dev queries & Test queries \\
\midrule
Math & MATH-500~\citep{lightman2024let} & 25 & 3 & 17 \\
     & AMC~\citep{aimo2024amc} & 4 & 1 & 1 \\
     & AIME 2024~\citep{maxwelljia2024aime} & 1 & 0 & 1 \\
\midrule
Code & LiveCodeBench v5~\citep{jain2025livecodebench} & 10 & 5 & 6 \\
     & LiveCodeBench v6~\citep{jain2025livecodebench} & 13 & 2 & 7 \\
     & MBPP~\citep{austin2021program} & 12 & 3 & 11 \\
\midrule
General reasoning & BBH~\citep{suzgun2023challenging} & 36 & 10 & 24 \\
Abstract reasoning & ARC~\citep{chollet2019measure} & 19 & 6 & 14 \\
Logical reasoning & LSAT-AR~\citep{zhong2024agieval} & 30 & 0 & 30 \\
Medical QA & MedMCQA~\citep{pal2022medmcqa} & 30 & 0 & 30 \\
Document QA & SQuAD~\citep{rajpurkar2016squad} & 34 & 11 & 25 \\
Long-context QA & InfiniteBench LongBook~\citep{zhang2024bench} & 15 & 4 & 10 \\
\midrule
\textbf{Total} & & \textbf{229} & \textbf{45} & \textbf{176} \\
\bottomrule
\end{tabular}
\end{table}

\paragraph{Code verification.}
MBPP and LiveCodeBench queries include public test cases
in their problem statements:
three per query for MBPP
and two to four for LiveCodeBench.
When PoT is selected,
these public tests provide execution feedback
to guide test-time search.
For final evaluation,
we augment the public tests with 12 additional cases per query which are withheld from public test cases,
drawn from MBPP+ for MBPP
and from the hidden test suite for LiveCodeBench.
A solution receives a utility score of one
only if it passes every test in the combined suite,
and zero otherwise.
This separation provides a stricter assessment of code correctness and reduces the risk of overfitting to public test examples.

\paragraph{Document-based tasks.} 
These tasks evaluate the acquisition and reuse of document knowledge as parametric memory.
A motivating workload involves multiple queries about the same reference material,
such as a book, a research paper, or technical documentation.
A reusable adaptation can retain knowledge beyond the current context
and support subsequent queries,
creating an opportunity to amortize adaptation cost
across repeated use.
We therefore separate document ingestion from question answering
and omit the original text from the answer-generation context.
This provides a controlled setting
for evaluating whether the policy can construct useful document skills and select them for later questions.
For SQuAD, 
this follows the knowledge-incorporation setting of~\citet{zweiger2025seal}.
The selected SQuAD subset contains 70 queries associated with 31 passages,
averaging 2.26 queries per passage.
LongBook extends this evaluation to longer reference material,
containing 29 queries associated with 25 books,
averaging 1.16 queries per book.
We partition these sources into training, development and test splits at the document level,
keeping each passage or book and all its associated queries in the same split.
Neither document identifiers nor document texts overlap across splits.
This separation tests generalization to unseen documents
and prevents direct memorization
of document-specific routing sequences
for the evaluation documents during policy training.
Each document defines a sequence
beginning with an unscored ingestion event,
followed by its associated questions.
During RL training,
we randomize the interleaving of groups
while preserving the order within each group.
This ensures that each document arrives
before any of its associated questions
and preserves the accumulation-before-reuse order
of the TTRL sequences.
The same ordering constraints apply to evaluation.
The evaluation stream therefore contains 19 document events (11 passages and 8 books) 
in addition to the 176 scored queries,
giving 195 events in total.
During document ingestion,
the policy can archive material
and construct adaptations for subsequent questions.
These document events receive no task-accuracy score and are excluded from the utility denominator;
overall utility is computed over the 176 scored queries.
Any adaptation performed during document ingestion 
still incurs computational cost.

\subsection{Benchmark Scope and Implications}

\paragraph{Distributional bias and evaluation scope.}
Queries with observed adaptation benefits constitute
a majority of both our training and test splits,
accounting for approximately 65\% of the curated benchmark overall.
This composition deliberately differs from the source distributions.
For comparison,
only 35 of the 500 queries in the complete MATH-500 dataset
(7.0\%) satisfy our high-gain criterion
under the PoT reference procedure with Qwen2.5-7B-Instruct.
The prevalence of beneficial adaptation depends on
the backbone, data, TTT method, and computation budget.
We deliberately enrich both training and evaluation
with beneficial cases
to provide substantial improvement opportunities
and examine how effectively TTT agency can exploit them.
The resulting gains therefore characterize this curated setting
and should not be interpreted as expected gains
under the original source distributions.

\paragraph{Influence on adaptation behavior.}
The prevalence of beneficial queries
also shapes the policy's learning incentives.
Our training distribution frequently rewards
appropriate adaptation and skill reuse,
encouraging more frequent use of TTT.
Under a distribution with fewer beneficial opportunities,
the same cost-sensitive objective favors more conservative behavior,
concentrating TTT on queries
for which adaptation offers positive expected net returns.
The desired behavior remains conditional:
the policy should learn where adaptation is worthwhile
from the outcomes observed during training.
Our pipeline prescribes no fixed adaptation frequency;
the learned behavior is shaped by
the observed trade-off between adaptation benefits and costs
under the training distribution.

\paragraph{Verifiability of TTT benefits.}
Our benchmark prioritizes tasks with explicit reference answers
or executable tests.
Fixed scoring rules reduce ambiguity in measuring adaptation gains
and support reliable utility estimates
without relying on LLM judges.
At deployment, reference answers are generally unavailable,
and task-provided checks can verify outcomes
only within their coverage.
Without ground truth or a reliable verifier,
we cannot directly confirm whether TTT helped
on an individual query.
The policy therefore draws on training experience
to select actions with positive expected returns
on similar deployment queries.
This is a general challenge for TTT without outcome feedback,
and resolving it falls outside the scope of this work.

\paragraph{Applicability to frontier models.}
Given constraints on access to model weights and training resources,
our benchmark targets the evaluated backbones
and is not designed to assess TTT agency in frontier models.
Evaluating TTT agency in such models calls for
substantially harder benchmarks at their capability boundaries,
where direct inference leaves meaningful room for improvement.
The underlying decision problem remains relevant:
whenever TTT can resolve otherwise unsuccessful tasks
at an acceptable cost,
there is value in learning to identify those tasks
and select an effective adaptation procedure.
Extending this evaluation therefore requires
new measurements on workloads tailored to the target model.
Such workloads could include challenging mathematical,
scientific, and engineering tasks,
potentially extending to open research problems
with independently verifiable outcomes.
The benchmark must evolve with model capabilities,
while the objective of learning conditional TTT decisions
remains applicable.

\section{Details of Policy Training}
\label{sec:app:training}

\paragraph{Training and evaluation settings.}
Unless otherwise specified,
we use Qwen2.5-7B-Instruct~\citep{hui2024qwen2}
with a rank-\(16\) policy LoRA.
Both training stages optimize only the policy LoRA using AdamW,
while keeping the backbone frozen.
Guidance distillation runs for two epochs
with learning rate \(10^{-5}\)
and batch size \(32\).
Reasoning and action tokens receive loss weights
\(\omega_z=1\) and \(\omega_a=4\), respectively,
and the summed weighted loss is divided by \(64\)
to set the gradient scale.
Reinforcement learning starts from the distilled checkpoint
and runs for three epochs
with learning rate \(5\times10^{-6}\).
Each rollout group contains \(G=8\) trajectories
sampled at temperature \(0.9\).
We use a GRPO clipping parameter of \(0.2\),
a KL coefficient of \(0.02\),
and gradient accumulation over three rollout groups.
An event is retired after all trajectories in one group
reach its measured reward bound,
with revisit probability \(0.05\).
The default cost per successful adaptation is \(0.1\),
and the invalid-action penalty coefficient is \(0.15\).
At evaluation time,
both policy generation and final-answer decoding are greedy.
For trained policies,
we report the mean and sample standard deviation
over three independent training runs,
evaluating each checkpoint once.

\paragraph{TTT methods.}
We equip Agentic-TTT with five TTT methods
covering different adaptation signals and temporal granularities.
\textsc{PoT}~\citep{jiao2026pot}
performs per-query search and online adaptation
using execution feedback or answer agreement.
\textsc{TTT-FS}~\citep{akyurek2025fsl}
converts in-context demonstrations into a task-specific adapter
through leave-one-out augmentation.
\textsc{TTRL}~\citep{zuo2025ttrl}
performs reinforcement learning over accumulated related queries
using majority-vote pseudo-labels.
\textsc{SEAL}~\citep{zweiger2025seal}
generates self-edits from input text
and distills them into an adapter.
\textsc{TLM}~\citep{hu2025tlm}
directly minimizes language-modeling loss on the input text.
Table~\ref{tab:ttt_execution}
summarizes their required materials,
training signals,
and hyperparameter settings.

\paragraph{Caching TTT outcomes.}
TTT adaptation is computationally expensive,
and the same adaptation configuration
may recur across training epochs.
We therefore cache its measured outcomes.
For each newly encountered configuration,
we perform three independent TTT runs,
evaluate the resulting adapters,
and cache their mean utility gain
as an estimate of the expected gain.
Subsequent occurrences reuse this estimate
instead of repeating the adaptation.
This reduces training overhead
and mitigates reward variability
caused by stochastic TTT optimization.

\paragraph{Training cost.}
We train on a single NVIDIA H200 GPU
and use vLLM to accelerate inference wherever supported.
Without TTT caching,
a training run with a 7B backbone
takes approximately \(11\) GPU-hours.

\section{Details of Execution Harness}
\label{sec:app:harness}

\subsection{Policy Interaction and Execution}

\paragraph{Policy reasoning and answer generation.}
The harness maintains separate contexts for policy interaction
and answer generation.
The policy context contains the current query,
a relevance-ranked view of the adaptation pools and available skills,
and the interaction history accumulated within the current query.
At each turn,
the frozen backbone equipped with the learned policy LoRA
generates a reasoning trace followed by a structured action.
The harness executes the action
and, for a nonterminal action,
appends its feedback and an updated state snapshot
to the conversation.
The policy then continues from this expanded context,
retaining its previous reasoning,
actions,
and observations.
For final-answer decoding,
the harness disables the policy LoRA
and uses the frozen backbone either without an adapter or with the selected skill LoRA,
as specified by the action sequence.
The answer-generation context contains only the original query;
it excludes the policy reasoning,
action history,
and environment observations.
The resulting response is submitted as the final answer,
terminating the interaction for that query.

\paragraph{Action protocol and execution constraints.}
Each policy turn emits one structured action
following its reasoning trace.
Table~\ref{tab:harness_actions}
summarizes the action arguments,
execution effects,
and preconditions.
Only \textsc{Generate} terminates the interaction;
the other actions return observations
that inform the policy's next decision.
Each query allows at most six policy turns
and one successful adaptation.
Parsing errors,
unmet preconditions,
and recoverable execution failures
are reported through observations,
allowing the policy to revise its action.
Every policy turn consumes the turn budget,
including rejected actions and policy-issued retries,
whereas a rejected \textsc{Adapt}
does not consume the adaptation allowance.
If the turn budget is exhausted
before the policy submits an answer,
the harness forces direct answer generation
with the frozen backbone and no skill adapter.

\begin{table}[h]
    \centering
    \small
    \setlength{\tabcolsep}{4pt}
    \renewcommand{\arraystretch}{1.15}
    \caption{
        Harness action interface.
    }
    \label{tab:harness_actions}
    \begin{tabular}{
        @{}
        p{0.11\linewidth}
        p{0.22\linewidth}
        p{0.31\linewidth}
        p{0.29\linewidth}
        @{}
    }
        \toprule
        Action
        & Arguments
        & Effect
        & Preconditions \\
        \midrule

        \textsc{Assign}
        &
        \texttt{pool\_key};
        optional \texttt{payload}
        &
        Archives the current query and supplied material
        in the specified pool,
        creating it if necessary.
        &
        The pool key must be permitted for the query;
        referenced material must be available.
        \\[3pt]

        \textsc{Adapt}
        &
        \texttt{method},
        \texttt{pool\_key};
        method-specific arguments
        &
        Trains or updates the pool's skill adapter
        and makes it active.
        &
        The pool must exist,
        satisfy the method's input and sample requirements,
        and have an available adaptation allowance.
        \\[3pt]

        \textsc{Load}
        &
        \texttt{skill\_id}
        &
        Makes the specified adapter active,
        replacing any previously active adapter.
        &
        The identifier must refer to an existing
        skill adapter.
        \\[3pt]

        \textsc{Generate}
        &
        \texttt{use\_skill}
        &
        Submits the final answer
        and terminates the interaction.
        Uses the active skill if requested and available;
        otherwise uses the frozen backbone.
        &
        No prior adaptation or loading is required.
        \\

        \bottomrule
    \end{tabular}
\end{table}

\paragraph{TTT execution.}
The policy provides method-specific payloads
through structured actions.
The harness parses each \textsc{Adapt} action block,
retrieves the inputs from the specified pool,
and executes the requested TTT method.
Table~\ref{tab:ttt_execution}
summarizes the required materials,
training signals,
and hyperparameter settings.

\begin{table}[t]
    \centering
    \scriptsize
    \setlength{\tabcolsep}{3pt}
    \renewcommand{\arraystretch}{1.25}
    \caption{
        TTT execution configurations.
        LoRA settings report rank \(r\),
        scaling parameter \(\alpha\),
        and dropout \(d\).
        QV denotes the query and value projections;
        Attn denotes all four attention projections;
        MLP denotes the gate, up, and down projections.
    }
    \label{tab:ttt_execution}
    \begin{tabular}{
        @{}
        p{0.10\linewidth}
        p{0.20\linewidth}
        p{0.18\linewidth}
        p{0.32\linewidth}
        p{0.13\linewidth}
        @{}
    }
        \toprule
        Method
        & Required Material
        & Training signal
        & Hyperparameter settings
        & LoRA \((r,\alpha,d)\) \\
        \midrule

        PoT
        &
        Query text; \newline
        optional executable tests. 
        &
        Reinforcement learning from execution feedback or answer agreement.
        &
        Up to 8 search expansions,
        3 children per expansion,
        maximum depth 8;
        learning rate \(10^{-4}\),
        3 PPO epochs per update.
        &
        \((8,16,0)\);\newline
        Attn.
        \\
        \arrayrulecolor[gray]{0.85}\midrule
        \arrayrulecolor{black}

        TTT-FS
        &
        In-context demonstrations.
        &
        Supervised reconstruction
        through demonstration augmentation.
        &
        Up to 250 augmentation per query;
        2 training epochs,
        batch size 2,
        learning rate \(10^{-4}\).
        &
        \((128,16,0)\);\newline
        QV + MLP.
        \\
        \arrayrulecolor[gray]{0.85}\midrule
        \arrayrulecolor{black}

        TTRL
        &
        A pool of compatible queries.
        &
        Reinforcement learning from majority-vote pseudo-labels.
        &
        Minimum pool size 16;
        16 rollouts per query,
        8 retained for optimization;
        4 epochs,
        prompt batch size 4,
        learning rate \(10^{-4}\).
        &
        \((16,32,0.05)\);\newline
        Attn.
        \\
        \arrayrulecolor[gray]{0.85}\midrule
        \arrayrulecolor{black}

        SEAL
        &
        Input text.
        &
        Supervised learning
        from generated self-edits.
        &
        5 self-edits;
        10 training epochs,
        batch size 1,
        learning rate \(10^{-3}\).
        &
        \((32,64,0)\);\newline
        QV.
        \\
        \arrayrulecolor[gray]{0.85}\midrule
        \arrayrulecolor{black}

        TLM
        &
        Input text.
        &
        Perplexity-weighted
        next-token prediction.
        &
        One update per 512-token chunks;
        log-perplexity threshold \(-3\),
        learning rate \(2\times10^{-4}\).
        &
        \((16,32,0.05)\);\newline
        Attn + MLP.
        \\

        \bottomrule
    \end{tabular}
\end{table}

\subsection{Pools, Skills, and Retrieval}

\paragraph{Pool construction.}
The policy decides whether to archive the current query
based on its content,
the relevance of the visible pools,
and the material available for adaptation.
To perform \textsc{Assign},
it specifies a pool name and a structured payload.
The harness constructs pool names
from a query-type prefix
and a descriptor or identifier derived from the input.
For example,
\texttt{mcq-medical-4opt}
encodes the multiple-choice format,
a rule-derived domain label,
and the number of options,
The resulting name is shown to the policy
and validated when processing \textsc{Assign}.

\paragraph{Skill lifecycle and representation.}
Each pool retains its archived samples
and at most one current skill adapter.
The first successful \textsc{Adapt} creates this adapter.
For incremental re-adaptation,
the harness warm-starts from the existing adapter
and trains on newly accumulated samples
together with up to eight randomly selected replay samples.
The updated adapter replaces its predecessor,
while the archived samples remain in the pool.
Pools and stored adapters persist across queries.
Each visible pool entry shows its name,
material excerpts,
total and unconsumed sample counts,
and training state,
along with the adapter ID and TTT method when available.
The training state indicates whether the pool is
\texttt{untrained},
\texttt{up-to-date},
or \texttt{growing} with additional unconsumed samples.
A growing pool's existing adapter remains available for reuse
until the policy requests another adaptation.

\paragraph{Relevance-based retrieval.}
The harness ranks pools using semantic embeddings
and BM25 lexical matching over query text and material excerpts.
For each pool,
semantic similarities are aggregated over its most relevant samples
and rescaled to obtain the displayed relevance score.
The semantic and lexical rankings are then combined
to select the top-\(k\) pools presented to the policy,
limiting the context overhead as the skill library grows.
We set \(k=8\) in our experiments.

\subsection{Prompt Templates}

\paragraph{Policy and observation templates.}
The following system prompt specifies the policy's role,
the available actions and TTT methods,
and the required response format.
This is the system prompt we consistently used in our model training and the prompted baseline.

\begin{tcolorbox}[myframe=gray,title={Policy system prompt}]
\begin{Verbatim}[fontsize=\scriptsize,breaklines=true,breakanywhere=false,breaksymbolleft={}]
You are a test-time policy controller for test-time training (TTT): you decide when and how to train yourself -- the frozen base model -- at test time, so that each incoming query is answered as well as possible. Each query is shown to you as CURRENT QUERY in the user message, together with a snapshot of your SKILLS; you decide, turn by turn, how to produce its answer.

A Skill is a sample pool (queries filed into it) plus the LoRA adapter that TTT (the `adapt` action) trains on those samples. The adapter is created the first time the Skill is adapted; before that the Skill holds samples only.

The snapshot lists Skills most-relevant-first, each with a relevance score `rel` and a training state:
  - `untrained` -- samples but no adapter yet (adapt to create one).
  - `up-to-date` -- its adapter covers all its samples (loadable).
  - `growing` -- new samples have arrived since the last adapt; the adapter stays loadable, and the new samples fold in at the next re-adapt.
A trained Skill's line also names its loadable adapter id and method. Judge a Skill's fit for the current query from its name, its content stamps (materials / pooled queries), its score, and its state.

Growing vs starting Skills:
  - Each Skill has ONE adapter. Incremental re-adaptation is warm-started on the existing adapter using newly accumulated samples and replay samples. The updated adapter replaces the existing one.
  - If a TTT method needs structured input, pass it via `assign_to_pool`'s `payload`. To store the CURRENT query's own material, write the literal value "<VERBATIM_QUERY>" and the harness fills in the query's source material exactly: `"payload": {"demos": "<VERBATIM_QUERY>"}` for demonstrations, or `"payload": {"context": "<VERBATIM_QUERY>", "title": "<short title you write>"}` for context text. If the query carries no such material, the assign is refused with an explanation. Write payload content explicitly when storing a subset or something other than the query's own material.

Adapt floors: each batch method has ONE sample floor -- a FIRST adapt is refused until the Skill holds that many samples, and a RE-adapt is refused until that many NEW samples have accumulated since the last adapt. The numbers are in the method list below, and the Skill listing shows live progress. Below a floor, keep buffering related queries and answer meanwhile with what exists -- the Skill's already-trained adapter, or the base model.

pot also produces a candidate answer during adaptation: its search keeps the best attempt it found, and generation with that adapter returns the stored attempt for the same query.

Operating rules:
  - generate ENDS the query: its output is the final answer. Take any assign_to_pool / load / adapt actions before it; when nothing better remains, generate.
  - assign_to_pool only files the current query for a possible future adapt -- it does not answer the query.
  - If an adapt is refused, do not retry it unchanged; use the feedback to revise your next action.
  - You have at most 6 turns per query.
  - The Skill listing comes only from observations; never invent or restate it.

You can take these actions, one per turn:
  * assign_to_pool {"pool_key": str, "payload": {..}?}
    File the CURRENT query into the Skill named `pool_key` (created if new), so related queries accumulate for a future adapt. Optional `payload` carries structured material from the query. Does NOT answer the current query -- the query is answered only when you call generate.

  * adapt {"method": str, "pool_key": str, "vote_signal": str?, "options": [..]?, "pattern": str?}
    Run TTT `method` over Skill `pool_key` to train or update its adapter from its samples, then auto-load it. Some methods need extra args (see the method list). An infeasible adapt is REFUSED; use the refusal message to revise your next action.

  * load {"skill_id": str}
    Set the active adapter to a stored Skill's adapter by its id (arg keyed `skill_id`; single slot; may be called again to swap).

  * generate {"use_skill": bool}
    Produce the answer to the current query and END it -- this is the TERMINAL action; its output is submitted as the final answer. use_skill=true requests the active LoRA; false answers with the bare base. Take any assign_to_pool / load / adapt actions BEFORE generating.

TTT methods available to `adapt`:
  - pot [per_query]: Repeatedly drafts candidate solutions, evaluates them using execution feedback or answer agreement, and learns from that feedback before drafting again. Optional executable tests are supplied through payload {"tests": [...]}.

  - seal [per_query]: Generates self-edits from the supplied input text and trains a LoRA on the generated material. REQUIRES each query to carry `context`; an optional `title` accompanies the text.

  - tlm [per_query]: Trains a LoRA directly on pooled input text using next-token prediction, without answer labels. Chunking of long inputs is automatic.

  - ttrl [batch; floor 16]: Trains using reinforcement learning from majority-vote pseudo-labels generated over pooled queries. A first adapt needs at least 16 samples; a re-adapt needs at least 16 NEW samples since the last adapt. REQUIRES `vote_signal` specifying the answer extraction rule: numeric | multiple_choice | short_answer | regex. regex also requires `pattern`; multiple_choice may take `options`.

  - ttt_few_shot [per_query]: Builds a LoRA through leave-one-out augmentation of input-output demonstrations. REQUIRES `payload.demos`, supplied as [{"input": ..., "output": ...}, ...] or using "target" instead of "output" for text-answer tasks.

Format -- EVERY turn, no exceptions: first reason inside <think>...</think> -- weigh your options (can the base model answer this directly? is a listed Skill worth loading? does any TTT method fit, and on what input?) -- then end the turn with EXACTLY one line:
ACTION: {"action": "<name>", "args": { ... }}
This holds for every action on a query, not just the first: an ACTION line with no <think> block before it is a format error.
\end{Verbatim}
\end{tcolorbox}

\paragraph{TTT-specific prompts and input construction.}
The following templates specify the internal prompts
and training-input formats of the TTT methods.
Braced fields denote substituted content,
and role labels indicate message boundaries.
TTRL samples responses directly from the queries stored in the pool,
without adding a method-specific instruction.
TLM trains directly on the input text or its chunks,
without an additional generation prompt.

\begin{tcolorbox}[
    myframe=gray,
    title={SEAL: self-edit generation and training text}
]
\begin{Verbatim}[
    fontsize=\scriptsize,
    breaklines=true,
    breakanywhere=false,
    breaksymbolleft={}
]
[System]
You are an assistant tasked with analyzing the provided passage and
producing a list of implications derived directly or indirectly
from the content.

[User]
Passage:
{title}
{context}
\end{Verbatim}

\small
Generated self-edits are split into segments.
Each segment is prefixed with the title
to form a training sequence.
The original passage is also included as training text.

\begin{Verbatim}[
    fontsize=\scriptsize,
    breaklines=true,
    breakanywhere=false,
    breaksymbolleft={}
]
[Self-edit training sequence]
{title}
{self_edit_segment}

[Original-passage training sequence]
{title}
{context}
\end{Verbatim}
\end{tcolorbox}

\begin{tcolorbox}[
    myframe=gray,
    title={PoT: candidate generation and revision}
]
\small
Each search expansion uses a single user message.
The initial message contains the problem;
a revision message additionally includes the parent candidate
and its feedback.

\begin{Verbatim}[
    fontsize=\scriptsize,
    breaklines=true,
    breakanywhere=false,
    breaksymbolleft={}
]
[Code: initial user message]
You are an expert Python programmer. Solve the following problem.

{problem}

{test_instruction}

Think briefly, then give the final solution in a single ```python code block.

[Code: test_instruction when tests are displayed]
Your solution must be a complete, self-contained Python program
that passes these tests:
```python
{tests}
```

[Code: test_instruction when tests are not displayed]
Write a complete, self-contained solution.

[Code: revision user message]
{initial_code_message}

A previous attempt was:
```python
{previous_code}
```

Execution feedback:
{execution_feedback}

Fix the issues and give a corrected, complete solution in a single
```python code block.


[Math: initial user message]
{problem}

Please reason step by step, and put your final answer within \boxed{}.

[Math: revision user message]
{initial_math_message}

A previous attempt is shown below.
---
{previous_answer}
---
{feedback} Carefully re-derive the solution from scratch, check each
step, and put your final answer within \boxed{}.

[Math: feedback when an answer is extracted]
Independent attempts at this problem disagree, so this answer
(\boxed{{previous_answer}}) may be wrong.

[Math: feedback when no answer is extracted]
The attempt produced no \boxed{} answer.
\end{Verbatim}

\small
Code feedback reports test results or execution errors.
Mathematics uses answer agreement to assign rewards.
\end{tcolorbox}

\begin{tcolorbox}[
    myframe=gray,
    title={TTT-FS: demonstration-based training inputs}
]
\small
Training examples are constructed by holding out a demonstration
and using the remaining demonstrations as context.
The held-out output supplies the supervised target,
and prompt tokens are masked from the training loss.
Text demonstrations are reordered;
grid demonstrations additionally undergo augmentation.
Ellipses below denote repeated demonstration pairs.

\begin{Verbatim}[
    fontsize=\scriptsize,
    breaklines=true,
    breakanywhere=false,
    breaksymbolleft={}
]
[Text tasks: user message]
Solve the last problem. Follow the format of the examples and respond
with ONLY the final answer, no explanation.

{input_1}
Answer: {target_1}

{input_2}
Answer: {target_2}

...

{held_out_input}
Answer:

[Text tasks: assistant target]
{held_out_target}


[Grid tasks: system message]
Figure out the underlying transformation in the following examples
and apply it to the test case. Here are some examples from this
transformation, your answer must follow the format.
The input-output grids are provided as python arrays:
{input_grid_1} -> {output_grid_1}#
{input_grid_2} -> {output_grid_2}#
...

[Grid tasks: user message]
{held_out_input_grid} ->

[Grid tasks: assistant target]
{held_out_output_grid}#
\end{Verbatim}
\end{tcolorbox}

\paragraph{Hinted collection templates.}
During guided rollout collection,
the policy receives a hint specifying the next target action
and explaining its execution.
The first-turn hint is appended to the current query,
while subsequent hints are appended to the corresponding observations.
The first box provides the shared wrappers and action descriptions.
For the first turn,
the wrapper contains one rationale from the second box;
for compute-aware direct-generation examples,
a rationale from the third box replaces the regular rationale.
Later turns use the shorter wrapper,
with a state or cost reminder when applicable.
An exact action block accompanies the action description.
Target actions and guidance variants
are selected from the training annotations.
Query excerpts come from the current input;
pool names,
adapter identifiers,
and sample counts
are filled from the target action and current harness state.
All collection hints are removed from the policy input prefixes
when constructing SFT examples
and are absent during evaluation.

\begin{tcolorbox}[myframe=gray,title={Collection hints: shared format}]
\begin{Verbatim}[fontsize=\scriptsize,breaklines=true,breakanywhere=false,breaksymbolleft={}]
[First-turn wrapper]
<COLLECTION_HINT>: Here is the guidance for this query. <RATIONALE> In <think>, reason ONLY about the next test-time-training action and why it is appropriate; do NOT solve or answer the question itself, since answer generation is handled separately. Explain the target action in your own words using this guidance and the current observation. Do not quote the collection hint. Then <FIRST_ACTION_DESCRIPTION>.

[Subsequent-turn wrapper]
<COLLECTION_HINT>: Next, <NEXT_ACTION_DESCRIPTION>. <STATE_OR_COST_REMINDER>

[Action description: assign to a new pool]
File this query into a new pool named "<POOL_KEY>", carrying the required material in the payload.

[Action description: assign to an existing pool]
File this query into the existing pool "<POOL_KEY>", carrying any required material in the payload.

[Action description: adapt]
Train the pool "<POOL_KEY>" with method "<METHOD>".

[Action description: load]
Load the trained adapter "<SKILL_ID>".

[Action description: generate with a skill]
Answer using the active skill adapter.

[Action description: direct generation]
Answer directly with the base model.

[Exact-action suffix, appended to either wrapper]
Commit EXACTLY the following action, preserving its arguments and any material-reference placeholders. Do not add arguments:
ACTION: <ACTION_JSON>
\end{Verbatim}
\end{tcolorbox}

\begin{tcolorbox}[myframe=gray,title={Collection hints: regular guidance}]
\begin{Verbatim}[fontsize=\scriptsize,breaklines=true,breakanywhere=false,breaksymbolleft={}]
[BUILD: TTT-FS]
The query includes input-output demonstrations. The target sequence builds a skill with ttt_few_shot, which uses these demonstrations for leave-one-out training. The order of actions matters: FIRST assign the query with the demonstrations in payload {"demos": ...}, so the pool contains the training material. Only AFTER the assign do I adapt with ttt_few_shot on that pool, and then answer using the resulting skill.

[BUILD: TTRL]
Filing this query brings the pool to <N_AFTER_ASSIGN> samples, meeting TTRL's minimum of 16. TTRL trains over pooled queries using reinforcement learning from majority-vote pseudo-labels. The target sequence files this sample, adapts the pool with TTRL, and answers with the resulting skill.

[BUILD: PoT]
PoT searches over candidate solutions, obtains execution feedback or answer-agreement signals, and updates the model before generating further candidates. The target sequence files the query and any supplied executable tests, invokes PoT, and then generates the answer.

[REUSE: TTRL]
The pool already has a trained adapter, which remains available for loading as new samples accumulate. The target sequence has two steps before answering: FIRST file this query into the pool, THEN load its trained adapter. Filing the query does not itself retrain the adapter.

[ACCUMULATE]
After filing this query, the pool will contain <N_AFTER_ASSIGN> samples, below TTRL's minimum of 16. The target sequence archives the query and answers with the base model for now. Archiving retains material for possible later adaptation; it does not itself train a skill.

[State reminder: sample requirements]
A first TTRL adaptation requires at least 16 samples. Re-adaptation requires at least 16 NEW samples since the previous adaptation. Distinguish total samples from new samples when reading the observation. Adaptation is an explicit action and does not occur automatically when a threshold is reached.

[State reminder: loading]
A skill can be loaded because its trained adapter exists. Its sample count does not determine whether that existing adapter is loadable.

[Subsequent-turn format reminder]
Keep <think> to a few sentences about the next action. Do not solve the question itself. Close </think>, then emit the ACTION line exactly once.
\end{Verbatim}
\end{tcolorbox}

\begin{tcolorbox}[myframe=gray,title={Collection hints: compute-aware direct generation}]
\begin{Verbatim}[fontsize=\scriptsize,breaklines=true,breakanywhere=false,breaksymbolleft={}]
[DIRECT: no expected gain]
The query asks about: "<QUERY_EXCERPT>". For this collection example, adaptation is not expected to improve the answer. Training a new adapter would add computation without an expected benefit, so the target is to answer directly with the base model.

[DIRECT: marginal expected gain]
The query asks about: "<QUERY_EXCERPT>". Adaptation might provide a small improvement, but training an adapter incurs additional computation. Under the intended compute trade-off, the expected benefit across its anticipated uses does not justify that expense. The target is to answer directly with the base model.

[DIRECT: corresponding skill not built]
The query asks about: "<QUERY_EXCERPT>". The corresponding skill has not been built in the current trajectory. Building it now would incur additional training computation for limited expected benefit under the intended compute trade-off. The target is to answer directly with the base model.

[Replacement rationale when the target TTRL pool remains below its sample floor]
After filing this query, the pool will contain <N_AFTER_ASSIGN> samples, below the 16 required for TTRL. This adaptation is therefore unavailable at the current step. The target sequence retains the query in the pool and answers directly with the base model.

[Addition when the target sequence begins with assign]
Two steps: FIRST file this query into the specified pool, THEN answer directly with the base model, with no adaptation in between. Filing stores the query but does not perform TTT.

[Shared state reminder]
Describe the pools and adapters as shown in the current observation. Do not claim that an adapter is active, newly trained, or updated unless the action history supports that claim. Filing a query does not create or update an adapter. Do not deny the existence of skills shown in the observation.

[Subsequent-turn reminder before direct generation]
Answer with the base model using use_skill=false. If the relevant adaptation is below its sample floor, acknowledge that constraint. Otherwise, explain the expected benefit and computation trade-off supplied by the collection guidance. Do not infer that other visible skills are unusable merely because they were trained on different queries.
\end{Verbatim}
\end{tcolorbox}

\section{Limitations and Discussions}
\label{sec:app:limitations}

In this section, we discuss the limitations of this work
and highlight opportunities for future research.

\paragraph{Benchmark selection and distribution dependence.}
Our benchmark deliberately enriches queries
with observed TTT benefits
to evaluate how effectively the proposed pipeline
can train a policy to exploit adaptation opportunities.
This outcome-based selection limits
the generality of the reported gains.
On workloads where beneficial adaptation is rare
or yields only modest improvements,
the achievable aggregate gains may be substantially smaller,
and the performance gap over competing baselines
may narrow or disappear.
Our results therefore do not establish
a consistent advantage across arbitrary deployment distributions.
The potential benefit of TTT agency is constrained by
the available TTT methods,
their suitability for the target workload,
the backbone's remaining improvement headroom,
and the computation budget.
Policy learning can help realize this potential
through appropriate adaptation and reuse decisions,
but cannot guarantee gains where no TTT method can provide useful improvement.

\paragraph{Cross-query credit assignment.}
Our RL stage optimizes query-level returns,
while the training stream advances through annotated gold actions.
Equation~\eqref{eq:stream_objective} therefore specifies
the deployment objective,
which is not directly optimized over complete streams during training.
Sample accumulation and skill construction for future reuse
are primarily supported by the priors established through
heuristic-guided SFT,
while RL refines decisions using current-query outcomes.
Although RL rollouts are sampled from the policy,
their initial library and pool states
are generated by gold action histories.
This creates a mismatch with autonomous deployment,
where earlier policy decisions determine subsequent states,
and can introduce exposure bias.
Learning across queries requires addressing delayed credit assignment:
the benefit of collecting material or constructing a skill
may emerge much later,
and a single successful answer may depend on
several earlier decisions.
Attributing such benefits to the contributing actions
is a central challenge.
We leave training on autonomous streams
with explicit cross-query credit assignment to future work.
Such training could improve both long-term utility
and robustness to states arising from the policy's own decisions.

\paragraph{Safety and reversibility.}
TTRL's model-generated pseudo-labels may reinforce incorrect predictions,
and a shared skill library may propagate errors
or become a target for poisoning through adaptation material.
Our design provides a degree of containment:
each skill is an isolated LoRA adapter,
and the pretrained backbone remains frozen.
A harmful skill can therefore be disabled or removed independently,
without reverting the backbone or unrelated adapters.
Compared with continual full-parameter updates,
this separation makes harmful parameter changes
easier to isolate and roll back.
However, reversibility does not prevent harmful outputs
while a compromised skill is active
or undo their downstream consequences.
Detecting poisoned skills and assessing robustness
under adversarial adaptation remain important directions
for future work.

\end{document}